%% file: aaai25.tex
\documentclass[letterpaper]{article} 

\usepackage{aaai25} 
\usepackage{algorithm,algorithmicx,algpseudocode}
\usepackage{times}  
\usepackage{epsfig}
\usepackage{float}
\usepackage{amssymb}
\usepackage{bm,xspace}
\usepackage{verbatim}
\usepackage{multirow}
\usepackage{etoolbox,siunitx}
\usepackage{calc}
\usepackage{pifont,hologo}
\usepackage{dsfont}

\usepackage{helvet}  
\usepackage{courier}  
\usepackage[hyphens]{url}  
\usepackage{graphicx} 
\usepackage{natbib}  
\usepackage{caption} 
\usepackage{comment} 

\usepackage{newfloat}
\usepackage{listings}
\DeclareCaptionStyle{ruled}{labelfont=normalfont,labelsep=colon,strut=off} 
\usepackage{adjustbox}
\usepackage{multirow}
\usepackage{amsfonts}
\usepackage{amsmath}
\usepackage{booktabs}
\usepackage{footmisc}
\usepackage[utf8]{inputenc} 
\usepackage[T1]{fontenc}    
\usepackage{url}            
\usepackage{booktabs}       
\usepackage{nicefrac}       
\usepackage{microtype}      
\usepackage[table]{xcolor} 
\definecolor{lightblue}{rgb}{0.882, 0.913, 0.964}
\definecolor{midblue}{rgb}{0.706, 0.792, 0.902}
\definecolor{darkblue}{rgb}{0.564, 0.690, 0.850}
\definecolor{lightred}{rgb}{0.984, 0.905, 0.917}
\definecolor{superlightblue}{rgb}{0.878, 1.000, 1.000} 
\usepackage{colortbl}
\usepackage{hhline}
\usepackage{csquotes}
\usepackage{graphicx}
\usepackage{thmtools}
\usepackage{thm-restate}
\usepackage{mathtools}
\usepackage{amsthm}
\usepackage{afterpage}

\theoremstyle{plain}

\usepackage[english]{babel}
\usepackage{comment}
\usepackage{tabularx} 
\usepackage{makecell} 
\usepackage{textcomp}

\makeatletter 
\renewcommand{\fnum@algorithm}{\fname@algorithm} 
\makeatother  
\usepackage{relsize} 
\usepackage{enumitem} 

\def\MM{\mathbf{M}}

\def\lL{\mathcal{L}}

\def\btheta{{\bm\theta}}

\DeclareMathOperator*{\argmin}{arg\,min}

\DeclareMathSymbol{@}{\mathord}{letters}{"3B}

\def\latex/{\LaTeX}
\def\bibtex/{\hologo{BibTeX}}

\floatstyle{ruled}
\newfloat{listing}{tb}{lst}{}
\floatname{listing}{Listing}
\title{Multi-Scale Graph Learning for Anti-Sparse Downscaling}
\author{
    Yingda Fan\textsuperscript{\rm 1}, Runlong Yu\textsuperscript{\rm 1}, Janet R. Barclay\textsuperscript{\rm 2}, Alison P. Appling\textsuperscript{\rm 3}, Yiming Sun\textsuperscript{\rm 1}, \\Yiqun Xie\textsuperscript{\rm 4}, Xiaowei Jia\textsuperscript{\rm 1}
    }
\affiliations{
    \textsuperscript{\rm 1}Department of Computer Science, University of Pittsburgh\\
    \textsuperscript{\rm 2}U.S. Geological Survey, New England Water Science Center\\
    \textsuperscript{\rm 3}U.S. Geological Survey, Water Mission Area\\
    \textsuperscript{\rm 4}Department of Geographical Sciences, University of Maryland\\
    \{yif47, ruy59, yis108, xiaowei\}@pitt.edu, \{jbarclay, aappling\}@usgs.gov, xie@umd.edu
}

\begin{document}
\hbadness=2000000000
\vbadness=2000000000
\hfuzz=100pt

\maketitle

\input{sections/00_abstract}
\input{sections/01_intro}

\input{sections/02_background}

\input{sections/03_section_multi-scale_graph_learning}

\input{sections/04_section_multi-scale_optimization}

\input{sections/05_experiments}

\input{sections/06_discussion}

\input{sections/07_acknowledgements}
\bibliography{aaai25}

\clearpage
\input{sections/09_appendix}


\end{document}

%% file: sections/00_abstract.tex
\begin{abstract}

    Water temperature can vary substantially even across short distances within the same sub-watershed. Accurate prediction of stream water temperature at fine spatial resolutions (i.e., fine scales, $\leq$ 1 km) enables precise interventions to maintain water quality and protect aquatic habitats. Although spatiotemporal models have made substantial progress in spatially coarse time series modeling, challenges persist in predicting at fine spatial scales due to the lack of data at that scale.   
    To address the problem of insufficient fine-scale data, we propose a Multi-Scale Graph Learning (MSGL) method. This method employs a multi-task learning framework where coarse-scale graph learning, bolstered by larger datasets, simultaneously enhances fine-scale graph learning. Although existing multi-scale or multi-resolution methods integrate data from different spatial scales, they often overlook the spatial correspondences across graph structures at various scales. To address this, our MSGL introduces an additional learning task, cross-scale interpolation learning, which leverages the hydrological connectedness of stream locations across coarse- and fine-scale graphs to establish cross-scale connections, thereby enhancing overall model performance. 
    Furthermore, we have broken free from the mindset that multi-scale learning is limited to synchronous training by proposing an Asynchronous Multi-Scale Graph Learning method (ASYNC-MSGL).
    Extensive experiments demonstrate the state-of-the-art performance of our method for anti-sparse downscaling of daily stream temperatures in the Delaware River Basin, USA, highlighting its potential utility for water resources monitoring and management.


\end{abstract}

%% file: sections/01_intro.tex
\section{Introduction}
\label{section:introduction}



Predicting water temperature at fine spatial scale (e.g., for discrete stream reaches of 1 km or shorter) can enhance understanding of and planning for the effects of weather and climate changes on fish habitats, as stream water temperatures can vary greatly along stream reaches due to differences in shade, groundwater inflows, and the presence of deep and shallow sections. Stream temperature downscaling aims to predict water temperatures at a fine spatial scale, using insights partly derived from spatially coarse data (e.g., stream reaches of roughly 10 km long) at the same temporal resolution. In this paper, we use both ``scale" and ``resolution" to describe the density of reach segments modeled per unit area. Our evaluation focuses on daily predictions for the Delaware River Basin, which supplies drinking water to over 15 million people~\cite{williamson2015summary} and requires maintaining cool water temperatures to preserve the habitat of aquatic life~\cite{brett1971energetic, ravindranath2016environmental}.

Spatiotemporal graphs have shown utility in modeling stream networks~\cite{moshe2020hydronets,Jia2021PhysicsGuidedRecurrent,sun2021explore,chen2022physics,topp2023stream,jia2023physics,he2024fair,jia2021physics_simlr,chen2023physics} as they concisely capture the interactions amongst different stream segments. 
However, existing graph-based stream models are focused on simulating water temperature at a coarse resolution due to data limitations at finer resolutions. 
Specifically, collecting water temperature observations requires costly field visits for sensor installation and maintenance, thus limiting the spatial resolution of observations. As a result, existing water temperature datasets in most places can only support coarse-resolution models. These models are unable to reflect the local variations in water temperature at a fine resolution~\cite{Terry2022StreamTemperatureData} and thus remain limited in informing local management decisions. 

To work around sparsity in fine-scale observations, various techniques leverage coarse-resolution data to inform fine-scale predictions. Multi-resolution learning models, particularly those employing Gaussian processes, effectively integrate data across different scales~\cite{Yousefi2019aggregate, Hamelijnck2019aggregate}. Bayesian deconditioning refines low-resolution spatial fields by incorporating high-resolution information~\cite{Chau2021sta}. Techniques like Generative Adversarial Networks (GANs) and quantile perturbation have been employed for meteorological downscaling~\cite{Chaudhuri2020sta, Tabari2021sta}. Further, efficient algorithms have been developed to enhance learning accuracy from imprecise, coarse-labeled data~\cite{Fotakis2021CoarseLabels}, and multiple instance learning has been extended to fine-scale predictions from aggregate observations~\cite{zhang2020aggregate}. The combination of consistent supervision and residual feature augmentation has shown promise in advancing super-resolution prediction tasks~\cite{Qin2020SuperResolution}. Additionally, multi-resolution data fusion techniques are being used to improve high-resolution land cover mapping~\cite{Robinson2019LandCoverMapping}.

Although existing methods, such as multi-resolution learning, provide valuable frameworks for downscaling and fine-scale prediction, they are primarily designed for gridded data, such as images, and are less suitable for irregularly structured data, such as graphs. 
Hence, these approaches are not well-adapted to stream temperature prediction models that rely on spatiotemporal graph networks. Unlike these traditional methods, our approach, termed Multi-Scale Graph Learning (MSGL), emphasizes the topological logic inherent in graph networks. 
In graph networks, graphs at different scales represent the same physical entities (e.g., stream reaches). We can thus use a cross-scale distance matrix to describe the physical distances and positional relationships between pairs of stream reaches at coarse and fine scales. The proposed MSGL method aims to adaptively learn the dynamic behaviors of nodes and their interactions across different scales.

In detail, 
MSGL processes data simultaneously at different scales by considering three key tasks. The first task, \textbf{Coarse-scale graph learning (CSL)}, aims to extract patterns of water dynamics at the coarse scale using features $\mathbf{X}_{c}$ and labels $\mathbf{Y}_{c}$ derived from the coarse-scale dataset $\mathcal{D}_{\text{coarse}}$. The second task, \textbf{Cross-scale interpolation learning (CrSL)}, is designed to capture the relationships between stream segments at different scales. It takes coarse-scale input features $\mathbf{X}_{c}$ from the dataset $\mathcal{D}_{\text{coarse}}$ and utilizes an interpolation module to generate fine-scale predictions $\mathbf{\hat{Y}}_{f}$. The third task, \textbf{Fine-scale graph learning (FSL)}, focuses on predicting the target variable at the fine scale. This task integrates multi-scale information by combining fine-scale features $\mathbf{X}_{f}$ and latent representations derived from the cross-scale interpolation. 
The three tasks share a subset of parameters to transform the same set of input features into latent representations, which helps capture water dynamics processes influenced by these features. 
Inspired by existing optimization algorithms~\cite{Ozan2018MultiTaskLearning,Mahapatra2020ParetoMTL},
MSGL uses a Multi-Scale Optimization (MSO) algorithm to balance the contributions of three learning tasks in the training process.

We have explored both synchronous and asynchronous MSGL. Most multi-resolution methods adopt synchronous strategies, such as combining loss functions~\cite{Robinson2019LandCoverMapping}. A few asynchronous methods begin with pre-training on low-resolution datasets before advancing to multi-resolution training and fine-tuning~\cite{Ji2021SuperResolution}.
Our work differs from these by fully capitalizing on directional heat transport and spatially correlated heat sources within watersheds~\cite{Ward1985ThermalCharacteristics}, which ensure that coarse-scale temperatures are predictive of the mean temperature across fine-scale stream reaches in the same area. Thus, predictions from a coarse-scale model can be remapped to the fine scale to provide simulation labels for pre-training in MSGL, to be followed by fine-tuning on observed but sparse fine-scale data.
We refer to this approach as Asynchronous Multi-Scale Graph Learning (ASYNC-MSGL).

We conduct extensive evaluations on real-world watersheds by comparing MSGL and ASYNC-MSGL with other spatiotemporal graph baselines that either focus on fine-scale data or multiple scales. We also conduct ablation experiments to analyze the contributions of learning tasks at different scales to the downscaling process. The results highlight the method's superior predictive accuracy and robustness, particularly where data are sparse.

%% file: sections/02_background.tex
\section{Problem Definition and Background}
\label{section:background}

\subsection{Graph learning for predicting water temperature}
Stream networks consist of multiple interconnected stream segments through which water flows from upstream to downstream. Graphs have been widely used to represent  complex interactions among multiple segments in stream networks~\cite{moshe2020hydronets,Jia2021PhysicsGuidedRecurrent,sun2021explore,chen2022physics,chen2021heterogeneous, topp2023stream}. 
A stream network can be represented by a  graph $\{\mathcal{V},\mathcal{E},\mathbf{A}\}$, where the node set $\mathcal{V}$ contains the set of river segments, 
the edge set $\mathcal{E}$ contains
the edges between upstream and downstream river segments, 
and the adjacency matrix  $\mathbf{A}$ measures the connection strength between each pair of nodes. 
The graph-based methods have shown good performance in predicting water temperature~\cite{Jia2021PhysicsGuidedRecurrent,topp2023stream}, streamflow~\cite{moshe2020hydronets,sun2021explore}, and other water properties~\cite{li2024real}.

These methods remain limited in addressing 
sparse and imbalanced observations over different segments due to the substantial cost needed for measuring water properties \cite{Harmel2023WQCosts}. Moreover, the stream networks can be created at different spatial scales based on varying standards for splitting segments. When modeling fine-scale networks, the model needs to capture both local and long-distance dependencies over the graph, which remains challenging for most graph neural network models~\cite{di2023over}.  

\subsection{Definition of anti-sparse downscaling}
The objective of this work is to predict the target variable (daily water temperature) $\mathbf{Y}_f$ in different stream segments at fine spatial resolution. Our method is anti-sparse in the sense that it maintains reasonable downscaling performance even when the local labeled data are extremely sparse.   
We have input drivers for $N$ fine-resolution river segments over $T$ dates. We also have an adjacency matrix $\mathbf{A}_f$ containing the adjacency (inverse of stream distance) between each pair of upstream-downstream segments. In addition to the fine-resolution data, we have input drivers $\mathbf{X}_c$, the observed target variable $\mathbf{Y}_c$, and the adjacency matrix $\mathbf{A}_c$ over $M$ ($M$<$N$) coarse-resolution stream segments that cover the same watershed in less detail. 
We also use a cross-scale matrix $\mathbf{D}\in \mathbb{R}^{N\times M}$. Each entry \(\mathbf{D}_{ij}\) represents the inverse stream distance between each pair of fine and coarse segments. 
The input drivers $\mathbf{X}_f$ and $\mathbf{X}_c$ contain the same set of variables at the same temporal resolution (more details in dataset description below), but are just reported at different spatial resolutions. For example, coarse-resolution data in $\mathbf{X}_c$ are estimated as averages over larger polygons than the fine-resolution data in $\mathbf{X}_f$. Both coarse and fine observations ($\mathbf{Y}_c$ and $\mathbf{Y}_f$) are available only for certain segments on certain dates. Because observations are made at points and then aggregated to segments, we typically have a larger number of observations per segment and a greater observation density at the coarse resolution (Appendix B.1). The proposed work aims to leverage the spatiotemporal patterns learned from coarse-resolution data to aid in the prediction of $\mathbf{Y}_f$ at fine resolution. 


\subsection{Recurrent graph network (RGrN)}
As the base model for our work, we use RGrN, which couples temporal recurrence with spatial graph convolution.  
Previous studies have shown utility of this model in predicting stream water temperature and streamflow~\cite{Jia2021PhysicsGuidedRecurrent,topp2023stream,Barclay2023ProcessGuidedDL}. 
RGrN extends the standard LSTM structure by incorporating neighborhood information in state evolution. Specifically, it computes states as $\mathbf{s}^t = \mathbf{fg}^{t}\odot(\mathbf{s}^{t-1}+\mathbf{A}\cdot g(\mathbf{s}^{t-1}))+ \mathbf{ig}^{t}\odot \bar{\mathbf{s}}^{t}$, where $\mathbf{s}^t$ contains the states of all stream segments at time $t$, $\mathbf{A}$ is an adjacency matrix used to convolve states among nearby stream segments, $g(\cdot)$ is a transformation implemented by fully connected layers, $\odot$ is the entry-wise product, and $\mathbf{fg}$, $\mathbf{ig}$, and $\bar{\mathbf{s}}^t$ are the forget gate, input gate, and candidate cell state, respectively, as in the standard LSTM~\cite{hochreiter_long_1997}. The state $\mathbf{s}^t$ is used to generate the hidden representation $\mathbf{h}^t$ and predict the target variable.

\subsection{Multi-objective learning and optimization}
Multi-scale learning methods have been 
proposed to extract and fuse features at different scales~\cite{Jiang2020Deraining, Liu2021DeepFeatureEnsemble, Zhao2021FacialExpressionRecognition}. These methods have found success in many computer vision tasks, such as fine-grained visual categorization and image classification~\cite{He2017Multiscale3DCNN, Zhang2021Multibranch, Chen2021CrossViT,Gu2022HighResVisionTransformer}.  
One major issue in multi-scale learning is about optimizing and balancing the learning process across different scales. 
Building upon the proposed multi-scale learning framework, this work also introduces a 
multi-scale optimization algorithm (to be discussed later) to adjust the influence of the gradient for learning tasks at different scales.  
The proposed optimization algorithm extends the principles of the Pareto multi-objective optimization algorithm~\cite{Deb2005RobustPareto, Desideri2012MGDA, Ozan2018MultiTaskLearning, Mahapatra2020ParetoMTL, Suzuki2020MultiObjectiveBO} and the robust multi-objective optimization algorithm~\cite{Roberts2018RobustHybridPower, Zhou2018MultiObjectiveRobust, Daulton2022RobustMOBO}.

%% file: sections/03_section_multi-scale_graph_learning.tex
\section{Multi-Scale Graph Learning}
\label{section:multi-scale graph learning}


\begin{figure*}[!t] 
\centering
\includegraphics[width = 1\linewidth]{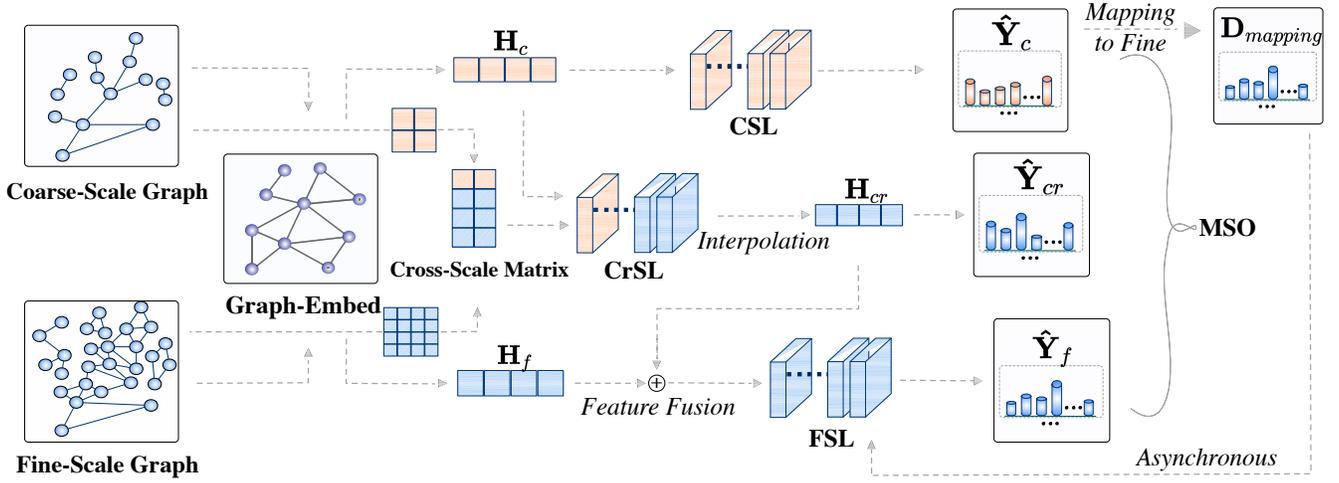}
\caption{The figure illustrates the architecture of the Multi-Scale Graph Learning (\textbf{MSGL}) model, which comprises three parallel training streams: coarse-scale graph learning (\textbf{CSL}), fine-scale graph learning (\textbf{FSL}), and cross-scale interpolation learning (\textbf{CrSL}). The trade-off among these three learning tasks is managed through Multi-Scale Optimization (\textbf{MSO}). $\mathbf{H}$ are hidden states. $\mathbf{\hat{Y}}$ are temperature predictions. 
$\mathbf{c}$, $\mathbf{cr}$, and $\mathbf{f}$ denote the coarse-scale, cross-scale, and fine-scale, respectively. $\mathbf{D}_{mapping}$ is a CSL-only $\mathbf{\hat{Y}}_{c}$ mapped to fine resolution for use in asynchronous pre-training of MSGL (\textbf{ASYNC-MSGL}).} 
\label{fig:architecture} 
\end{figure*}

As shown in Figure~\ref{fig:architecture}, the proposed framework contains two major modules, the shared spatiotemporal graph embedding module, abbreviated as \textbf{Graph-Embed}, and the scale-specific module. The shared graph embedding module converts input drivers at fine or coarse scale into latent representations: 
\begin{equation}\label{equation:graph-embed}
\mathbf{H} = \text{Graph-Embed}(\mathbf{X}, \mathbf{A};\Theta), 
\end{equation}
where \( \mathbf{H} \) represents the graph-embedded feature representation, encompassing comprehensive spatial and temporal information. The input \( \mathbf{X} \) and  \( \mathbf{A} \) can be from either the coarse or the fine spatial resolution (i.e., $\{\mathbf{X}_c,\mathbf{A}_c\}$ or $\{\mathbf{X}_f,\mathbf{A}_f\}$), and the parameters \(\Theta\) are shared across different spatial resolutions. The embedding function $\text{Graph-Embed}$ is implemented using 
the RGrN model~\cite{Jia2021PhysicsGuidedRecurrent}.  

The representation \( \mathbf{H}\) is then passed into the scale-specific prediction module, which conducts three predictive learning tasks to create predictions at different spatial scales. These tasks are designed to complement each other while being optimized simultaneously. Next, we describe the three learning tasks in the scale-specific module.   

\subsection{Synchronous multi-scale graph learning}
\label{section:scale-specific module}


\subsubsection{Coarse-scale graph learning. }
\label{subsection:task1}
The \textbf{CSL} task is designed to capture broad spatial patterns in water temperature dynamics from $\mathcal{D}_{\text{coarse}}$, with the expectation that fine-scale temperatures fluctuate around those coarse-scale patterns.

Given the input drivers $\mathbf{X}_c$ and adjacency matrix $\mathbf{A}_{c}$ at the coarse scale, we embed the hidden representations $\mathbf{H}_{c}$ through $\mathbf{H}_{c} = \text{Graph-Embed}(\mathbf{X}_{c},  \mathbf{A}_{c};\Theta)$.
The final output is generated through fully connected layers as  $\hat{\mathbf{{Y}}}_{{c}} = \text{Dense}(\mathbf{H}_{{c}};\Phi_c)$, 
where \(\hat{\mathbf{Y}}_{c}\) represents the predicted coarse-scale temperatures, and  \(\Phi_c=\{\mathbf{W}_c,\mathbf{b}_c\}\) represents the parameters for the fully connected layers that are specific to the CSL. 
In comparison to graphs relying solely on the fine-scale data $\mathcal{D}_{\text{fine}}$, 
the incorporation of CSL can help stabilize fine-scale predictions around coarse-scale trends, resulting in more robust performance. 

\subsubsection{Cross-scale interpolation learning. }\label{subsection:task2}
The \textbf{CrSL} task interpolates the coarse-scale graph embeddings $\mathbf{H}_c$ to generate fine-scale latent representations $\mathbf{H}_f$ that preserve the graph's structure and reflect the fine-scale distribution as 
\begin{equation}\label{eq:Interpolation}
\mathbf{H}_{{cr}} = \mathbf{D} \cdot \mathbf{H}_{c}, 
\end{equation}
where $\mathbf{D}\in \mathbb{R}^{N\times M}$ is the cross-scale matrix. 
Each entry \(\mathbf{D}_{ij}\) represents the adjacency (inverse of distance) between 
each pair of fine-scale node $i$ and coarse-scale node $j$. 



Next, the interpolated representation \(\mathbf{H}_{cr}\) is  processed through an attention mechanism~\cite{vaswani2017attention}, and then gets concatenated with the original representation \(\mathbf{H}_{cr}\), and finally normalized to produce the final feature representation \(\tilde{\mathbf{H}}_{cr}\), as $\tilde{\mathbf{H}}_{cr} = \text{BatchNorm}\left(\begin{bmatrix} \mathbf{H}_{cr}, \text{Multihead-Attn}(\mathbf{H}_{cr})\end{bmatrix}\right)$.  
The final output is generated through fully connected layers, as $\mathbf{\hat{Y}}_{cr} = \text{Dense}(\tilde{\textbf{H}}_{cr};\Phi_{cr})$, 
where $\Phi_{cr}$ represents the parameters for the fully connected layers that are specific to the cross-scale interpolation task. 
The task output $\hat{\mathbf{Y}}_{cr}$ is produced at the fine scale.

Equation~\ref{eq:Interpolation} integrates the cross-scale distance matrix $\mathbf{D}$ into the interpolation task within the implicit representation space, encoding cross-scale information by capturing relative node positions at different scales. This, along with an attention mechanism, enhances the model's understanding of intrinsic relationships between these cross-scale nodes, thereby improving overall performance



\subsubsection{Fine-scale graph learning. }\label{subsection:task3}
The \textbf{FSL}) task aims to directly predict high-resolution labels $\mathbf{Y}_f$ using $\mathbf{X}_f$ while also leveraging the representations $\mathbf{H}_{cr}$ from CrSL. 
Specifically, the high-resolution input \(\mathbf{X}_{f} \) is first embedded through the shared graph embedding module, as $\mathbf{H}_{f} = \text{Graph-Embed}(\mathbf{X}_{f},  \mathbf{A}_{f};\Theta)$. 
Although $\mathbf{H}_{f}$  has the same dimensions as $\mathbf{H}_{cr}$, it is important to note that $\mathbf{H}_{f}$  directly encodes the information from the fine-resolution inputs $\mathbf{X}_{f}$ and $\mathbf{A}_{f}$. 

To further enrich the representation, we combine the information obtained from both fine resolution and coarse resolution. In particular, 
concatenating the obtained $\mathbf{H}_{f}$ with the representation \( \mathbf{H}_{cr} \) of the cross-scale task forms a comprehensive feature representation \( \mathbf{H}_{f,cr} = \begin{bmatrix} \mathbf{H}_{f},  \mathbf{H}_{cr} \end{bmatrix} \). This integration enhances the combination of information from different scales. 
The combined features are then transformed through a fully connected layer, residual connection, and batch normalization, as $\tilde{\mathbf{H}}_{f} = \text{BatchNorm}\left(\mathbf{H}_{\text{f,cr}} + \sigma(\mathbf{W}_{\text{res}} \mathbf{H}_{\text{f,cr}} + \mathbf{b}_{\text{res}})\right)$. Then we create the final prediction as $\mathbf{\hat{Y}}_{f} = \text{Dense}(\tilde{\mathbf{H}}_{f};\mathbf{W}_{f},\mathbf{b}_f)
$, where $\sigma$ is the non-linear activation function and the parameters $\Phi_{f} = \{\mathbf{W}_\text{res}, \mathbf{b}_\text{res}, \mathbf{W}_{f}, \mathbf{b}_{h}\}$ are specific to \textbf{FSL}.

%% file: sections/04_section_multi-scale_optimization.tex
\subsubsection{Multi-scale optimization}
\label{section:multi-scale optimization}

\afterpage{
    \begin{figure*}[t]
        \centering
        \includegraphics[width=.8\linewidth] 
        {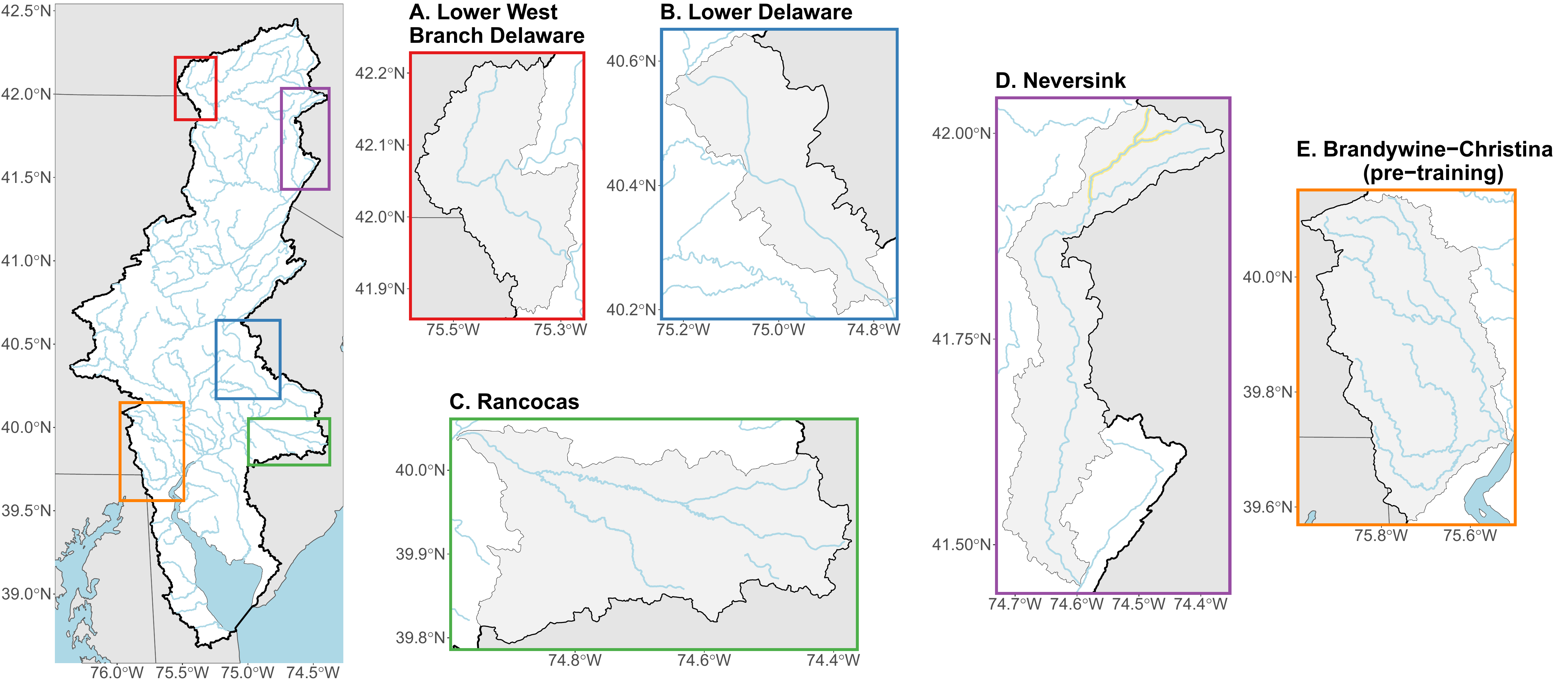}
        \caption{Map of five well-observed watersheds in the Delaware River Basin in the eastern United States. (Map layers from~\citealt{a_mullen_usaboundaries_2018},~\citealt{bock_geospatial_2020}, and~\citealt{WBD2023}.)}
        \label{fig:studyarea}
    \end{figure*}
}

Multi-scale optimization (\textbf{MSO}) trains the model by iteratively updating the scale-specific parameters $\Phi_\tau$ ($\tau\in\{c,cr,f\}$) and the shared parameters $\Theta$. 
We first update $\Phi_\tau$ using gradient descent. 
The update process is 
\begin{equation}
\Phi_{\tau} = \Phi_{\tau} - \eta \nabla_{\Phi_{\tau}} \mathcal{L}_{\tau}, 
\end{equation}
where \(\eta\) is the learning rate and \(\mathcal{L}_{\tau}\) is the loss function  for each task. Both $\mathcal{L}_{cr}$ and $\mathcal{L}_f$ are defined as the mean-squared loss on the fine-scale labels, and $\mathcal{L}_c$ is defined on the coarse-scale labels. The gradient \(\nabla_{\Phi_{\tau}}\) 
is calculated based on the loss \(\mathcal{L}_{\tau}\) with respect to the parameters \(\Phi_{\tau}\). 

We next update the shared parameters $\Theta$ using information from both scales. We denote by $\nabla_{\Theta} \mathcal{L}_{\tau}$, $\tau \in \{{c}, {cr}, {f}\}$, the gradient of the loss function of each scale-specific task with respect to the shared parameters $\Theta$. 
To optimize multi-scale learning tasks simultaneously without harming any particular task, 
we utilize the multiple-gradient descent algorithm (MGDA,~\citealt{Desideri2012MGDA}) 
to adjust the weight of loss gradient  for each learning task, as $\{\alpha_{\tau}\}$, for $\tau \in \{{c}, {cr}, {f}\}$ (detailed in Appendix A).
Finally, we update the shared parameters \(\Theta\) using the 
$\alpha_{\tau}$-weighted sum of gradients, as 
\begin{equation}
\Theta = \Theta - \eta \sum_\tau {\alpha}_\tau  {\nabla}_\Theta \mathcal{L}_\tau. 
\end{equation}




\subsection{Asynchronous multi-scale graph learning}\label{subsection: ASYNC-MSGL}
\subsubsection{CSL training, prediction and fine-scale mapping.}
We introduce an asynchronous MSGL (\textbf{ASYNC-MSGL}) method that begins with a standalone coarse-scale graph learning phase. Initially, the model is trained on abundant coarse-scale data collected over large regions using \textbf{CSL}. The trained model then generates predictions on the coarse-scale data, which are mapped to the fine scale, creating the labels for the fine-scale data in \( \mathcal{D}_{\text{mapping}} \).

To ensure consistency and continuity throughout the multi-task learning process during the \( \mathcal{D}_{\text{mapping}} \)-based pretraining, the coarse-scale labels in \( \mathcal{D}_{\text{mapping}} \) are not the original ground truth labels, but rather the predictions from the trained CSL model. The fine-scale labels are derived by remapping those CSL predictions from the coarse-scale reaches to the spatially corresponding fine-scale reaches.

\subsubsection{MSGL pre-training on \( \mathcal{D}_{\text{mapping}} \) and fine-tuning on \( \mathcal{D}_{\text{groundtruth}} \) \textbf{dataset.}}

Once the labels are generated, we proceed to pre-training on \( \mathcal{D}_{\text{mapping}} \), where the model learns from both coarse-scale and fine-scale inputs. Unlike the fixed periods in the ground truth dataset, \( \mathcal{D}_{\text{mapping}} \) covers a broader temporal range as its labels are derived from CSL predictions, not limited by observation data. This extended temporal coverage allows the model to capture longer-term dynamics and improve temporal continuity in predictions, enhancing its robustness against sparse data.

After pre-training, the model is fine-tuned on the \( \mathcal{D}_{\text{groundtruth}} \) dataset, which contains the original ground truth labels. This two-stage process—pre-training on simulated data followed by fine-tuning on real data—helps the model align with the true data distribution while maintaining the benefits of the broader scale learning.

%% file: sections/05_experiments.tex
\section{Experimental Results}
\label{section:experiments}

\paragraph*{Dataset.}

\newcolumntype{L}[1]{>{\raggedright\arraybackslash}p{#1}}
\newcolumntype{C}[1]{>{\centering\arraybackslash}p{#1}}
\newcolumntype{R}[1]{>{\raggedleft\arraybackslash}p{#1}}
\begin{table*}[ht]\scriptsize
    \centering
    \caption{Root mean squared errors (RMSE; mean and standard deviation of nine replicates) for various methods across four watersheds using 1\% random subsets of observed water temperatures for training. Single-scale baselines, including RGrN, STGCN, and DCRNN, as well as multi-scale baselines MS-STGCN and MSGNET, are referenced in RQ1-Baselines. Components of our method—CSL, CrSL, FSL, MSGL, and ASYNC-MSGL—comprise the ablation study described in RQ1-``Performance comparison." Bold values are the best results in each watershed and those not significantly worse (Welch's t-test, p > 0.05).}
    \label{tab:table1}
    \begin{tabular}{L{2.2cm}L{2.2cm}C{2.2cm}C{2.5cm}C{2.2cm}C{2.2cm}}
        \toprule
        \multirow{2}{*}{\textbf{Category}} 
        & \multirow{2}{*}{Method} 
        & \textsc{Lower Delaware} & \textsc{LwBranch Delaware} & \textsc{Neversink} & \textsc{Rancocas} \\
        & &{\textcolor{gray}{RMSE}}\phantom{$\pm$}{\textcolor{gray}{STDEV}} 
        &{\textcolor{gray}{RMSE}}\phantom{$\pm$}{\textcolor{gray}{STDEV}} 
        &{\textcolor{gray}{RMSE}}\phantom{$\pm$}{\textcolor{gray}{STDEV}} 
        &{\textcolor{gray}{RMSE}}\phantom{$\pm$}{\textcolor{gray}{STDEV}} \\
        \midrule
        \multirow{4}{*}{Single-scale Baselines}
        & STGCN (2018)
        & 1.653 \textcolor{gray}{$\pm$ 0.076} & 2.662 \textcolor{gray}{$\pm$ 0.107} & 2.059 \textcolor{gray}{$\pm$ 0.206} & 1.984 \textcolor{gray}{$\pm$ 0.130} \\
        & DCRNN (2018) 
        & 1.679 \textcolor{gray}{$\pm$ 0.040} & 2.487 \textcolor{gray}{$\pm$ 0.134} & 2.297 \textcolor{gray}{$\pm$ 0.120} & 2.489 \textcolor{gray}{$\pm$ 0.287} \\
        & GASTN (2020)
        & 2.141 \textcolor{gray}{$\pm$ 0.342} & 2.525 \textcolor{gray}{$\pm$ 0.118} & 2.331 \textcolor{gray}{$\pm$ 0.153} & 3.134 \textcolor{gray}{$\pm$ 0.658} \\ 
        & RGrN (2021)
        & 1.732 \textcolor{gray}{$\pm$ 0.131} & 2.648 \textcolor{gray}{$\pm$ 0.154} & 1.935 \textcolor{gray}{$\pm$ 0.090} & 2.198 \textcolor{gray}{$\pm$ 0.246} \\ 
        \midrule
        \multirow{2}{*}{Multi-scale Baselines}
        & MS-STGCN (2021)
        & 1.484 \textcolor{gray}{$\pm$ 0.117} & 2.402 \textcolor{gray}{$\pm$ 0.116} & 1.937 \textcolor{gray}{$\pm$ 0.069} & 1.723 \textcolor{gray}{$\pm$ 0.263} \\
        & MSGNET (2024)
        & 1.492 \textcolor{gray}{$\pm$ 0.118} & 2.313 \textcolor{gray}{$\pm$ 0.125} & 1.927 \textcolor{gray}{$\pm$ 0.060} & 1.813 \textcolor{gray}{$\pm$ 0.351} \\   
        \midrule
        \multirow{6}{*}{Ablation Study}
        & MSGL (CSL) 
        & 5.799 \textcolor{gray}{$\pm$ 2.171} & 5.465 \textcolor{gray}{$\pm$ 0.462} & 4.215 \textcolor{gray}{$\pm$ 0.150} & 5.965 \textcolor{gray}{$\pm$ 0.841} \\
        & MSGL (CrSL) 
        & 1.578 \textcolor{gray}{$\pm$ 0.112} & 2.791 \textcolor{gray}{$\pm$ 0.144} & 2.623 \textcolor{gray}{$\pm$ 0.293} & 1.863 \textcolor{gray}{$\pm$ 0.173} \\
        & MSGL (FSL)  
        & 1.732 \textcolor{gray}{$\pm$ 0.131} & 2.648 \textcolor{gray}{$\pm$ 0.154} & 1.935 \textcolor{gray}{$\pm$ 0.090} & 2.198 \textcolor{gray}{$\pm$ 0.246} \\
        & MSGL (w/o CSL)
        & 1.470 \textcolor{gray}{$\pm$ 0.056} & 2.391 \textcolor{gray}{$\pm$ 0.118} & 2.249 \textcolor{gray}{$\pm$ 0.290} & 1.814 \textcolor{gray}{$\pm$ 0.225} \\
        & MSGL (w/o CrSL) 
        & 1.551 \textcolor{gray}{$\pm$ 0.057} & 2.593 \textcolor{gray}{$\pm$ 0.167} & 2.048 \textcolor{gray}{$\pm$ 0.256} & 2.150 \textcolor{gray}{$\pm$ 0.183} \\
        & MSGL (w/o MSO) 
        & \textbf{1.391} \textcolor{gray}{$\pm$ 0.069} & 2.432 \textcolor{gray}{$\pm$ 0.108} & 2.035 \textcolor{gray}{$\pm$ 0.204} & 1.743 \textcolor{gray}{$\pm$ 0.201} \\    
        \midrule
        \multirow{2}{*}{Sync vs. Async}
        & MSGL 
        & \textbf{1.384} \textcolor{gray}{$\pm$ 0.082} & 2.375 \textcolor{gray}{$\pm$ 0.106} & 1.919 \textcolor{gray}{$\pm$ 0.058} & \textbf{1.450} \textcolor{gray}{$\pm$ 0.109} \\
        & ASYNC-MSGL
        & \textbf{1.360} \textcolor{gray}{$\pm$ 0.116} & \textbf{2.179} \textcolor{gray}{$\pm$ 0.129} & \textbf{1.617} \textcolor{gray}{$\pm$ 0.134} & \textbf{1.482} \textcolor{gray}{$\pm$ 0.053} \\
        \bottomrule
    \end{tabular}
\end{table*}

\begin{figure*}[ht]
\centering
\includegraphics[width=0.73\linewidth]{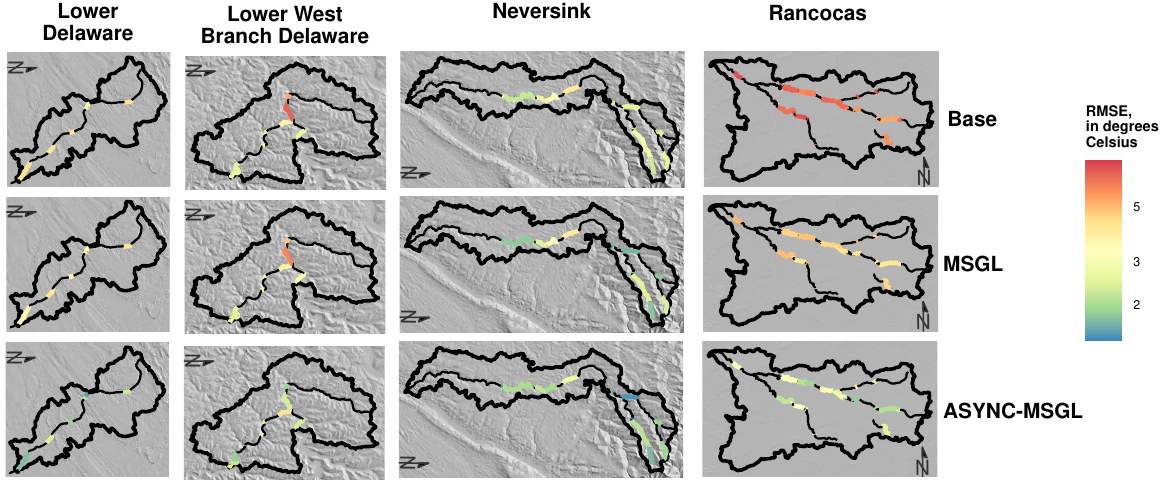}
\caption{Root mean squared error (RMSE, \textdegree C) of modeling methods RGrN (Base), MSGL, and ASYNC-MSGL applied to four watersheds, with 0.1\% of the fine-scale labels in each watershed used for training. Reaches of varying lengths are color-coded by average RMSE over all dates. (Map layers from~\citealt{EPA2012NHDPlus}.)}
\label{fig:spatialRMSE}
\end{figure*}

\begin{table*}[htbp]
\caption{Performance of Cross-Domain Pretraining (``MSGL (Pre-trained by Brandywine-Christina)") and Asynchronous MSGL (ASYNC-MSGL, i.e., ``MSGL (Pre-trained by $\mathcal{D}_{\text{mapping}}$)") in four watersheds. We simulated data sparsity by using 0.1\% or 1\% of fine-scale labels for fine-tuning. Dark blue indicates a large decrease (\(\Delta < -0.2\)), medium blue a moderate decrease (\(-0.2 \leq \Delta < -0.1\)), light blue a slight decrease (\(-0.1 \leq \Delta < 0\)), and light red any increase (\(\Delta > 0\)) in root mean squared error (RMSE, \textdegree C) relative to MSGL without Pre-training. Brandywine-Christina is one of the watersheds shown in Figure~\ref{fig:studyarea}.} 
\label{tab:pretrain}
\centering
\renewcommand{\arraystretch}{1.5} 
\scalebox{0.67}{
\begin{tabular}{|c|cc|cc|cc|cc|cc|cc|cc|cc|}
\toprule
\multirow{2}{*}{Model} & \multicolumn{4}{c|}{LowerDelaware} &  \multicolumn{4}{c|}{LwBranchDelaware} &  \multicolumn{4}{c|}{Neversink} &  \multicolumn{4}{c|}{Rancocas} \\
\cline{2-17}
                       & 0.1\% & \(\Delta\) & 1\% & \(\Delta\) & 0.1\% & \(\Delta\) & 1\% & \(\Delta\) & 0.1\% & \(\Delta\) & 1\% & \(\Delta\) & 0.1\% & \(\Delta\) & 1\% & \(\Delta\) \\

\midrule
MSGL (without Pre-training)        
& 2.521 & ---       & 1.384 & ---       & 2.687 & ---       & 2.375 & ---       & 2.323 & ---       & 1.919 & ---    & 4.431 & ---       & 1.450 & ---   \\
MSGL (Pre-trained by Brandywine-Christina)    
& \cellcolor{darkblue}2.006 & \cellcolor{darkblue}-0.515  & \cellcolor{lightred}1.408 & \cellcolor{lightred}0.024  & \cellcolor{lightred}2.936 & \cellcolor{lightred}0.249  & \cellcolor{lightred}2.495 & \cellcolor{lightred}0.12  & \cellcolor{lightred}2.664 & \cellcolor{lightred}0.341  & \cellcolor{midblue}1.759 & \cellcolor{midblue}-0.16  & \cellcolor{darkblue}3.855 & \cellcolor{darkblue}-0.576  & \cellcolor{lightred}1.489 & \cellcolor{lightred}0.039  \\
MSGL (Pre-trained by \( \mathcal{D}_{\text{mapping}} \))    
& \cellcolor{darkblue}1.818 &\cellcolor{darkblue} -0.703  & \cellcolor{lightblue}1.360 & \cellcolor{lightblue}-0.024  & \cellcolor{darkblue}2.376 & \cellcolor{darkblue}-0.311  & \cellcolor{midblue}2.179 & \cellcolor{midblue}-0.196  & \cellcolor{darkblue}2.036 & \cellcolor{darkblue}-0.287  & \cellcolor{darkblue}1.617 & \cellcolor{darkblue}-0.302  & \cellcolor{darkblue}2.794 & \cellcolor{darkblue}-1.637  & \cellcolor{lightred}1.483 & \cellcolor{lightred}0.033  \\
\bottomrule
\end{tabular}
}
\label{tab:RQ3}
\end{table*}

Figure~\ref{fig:studyarea} illustrates the study area, encompassing the Brandywine, Lower Delaware, Neversink, Lower West Branch Delaware, and Rancocas watersheds within the Delaware River Basin, eastern United States. The data
and predictions are daily temporal resolution, covering the years 1979-2021. Data were collected at both high and low spatial resolutions \cite{Oliver2022WaterQualityData,Terry2022StreamTemperatureData,gridmet2013}. Specifically, the coarse-scale river segments were defined by the geospatial fabric used for the National Hydrologic Model (NHM)
~\cite{Regan2018NationalHydrologicModel} and have an average segment length of 10.5 km in the Delaware River Basin. In contrast, the fine-scale segments were defined by the National Hydrography Dataset (NHD)~\cite{USGS2019NHD} and have an average segment length of 1.3 km. The input features include slope, elevation, width, daily average air temperature, solar radiation, precipitation, and potential evapotranspiration. 


Although the aforementioned watersheds have well-sampled fine-scale and coarse-scale data, fine-scale labels are very limited in most other watersheds. To more accurately evaluate our method in real-world downscaling scenarios, we conduct experiments using training and validation sets with intentionally sparsified fine-scale labels. Meanwhile, labels from testing sets remain intact to accurately assess the downscaling results. 
For details on dataset characteristics; the partitioning of training, validation, and testing sets based on years; and the use of label masking to simulate data sparsity, please refer to Appendix~\ref{appendix:details of data}.

\begin{figure*}[ht]
    \centering
    \begin{minipage}{1\textwidth}
        \centering
        \includegraphics[width=\linewidth]{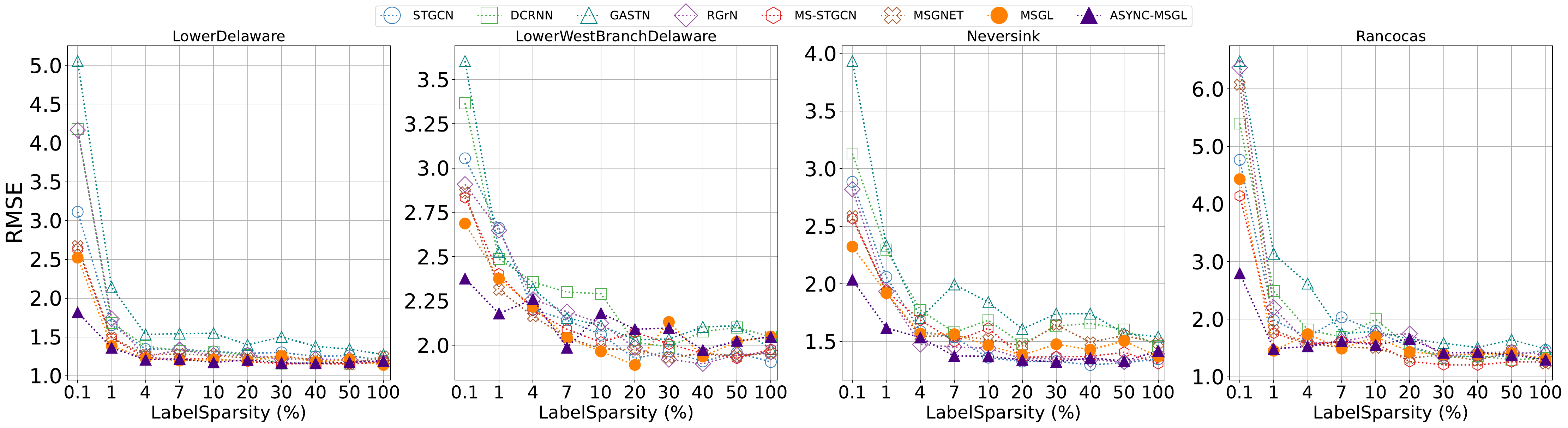}
    \end{minipage}
    \caption{Robustness of spatiotemporal graph models across varying label sparsity levels (training on 0.1\%, 1\%, 4\%, 7\%, 10\%, 20\%, 30\%, 40\%, 50\%, and 100\% of all fine-scale reach-days) in four watersheds. Filled markers represent our new methods MSGL and ASYNC-MSGL, whereas open markers represent the methods referenced in section RQ1-Baselines.}
    \label{fig:sparseRMSE}
\end{figure*}

\paragraph*{Research questions.}
We designed evaluations focusing on three core questions:\\
\textbf{RQ1.}   How well does MSGL downscale stream temperature compared to baseline models?\\
\textbf{RQ2.}   How robust is the MSGL model under varying levels of data sparsity?\\
\textbf{RQ3.}   Which is more effective: Cross-Domain Pretraining or Asynchronous MSGL?\\


\subsection{Downscaling performance \textbf{(RQ1)}}
\label{subsection:experiment1}

\paragraph*{Baselines.}

We evaluate MSGL against several baselines, including commonly used single-scale spatiotemporal graph models such as RGrN~\cite{Jia2021PhysicsGuidedRecurrent}, Spatio-Temporal Graph Convolutional Network (STGCN; \citealt{Yu2018STGCNAI}), Diffusion Convolutional Recurrent Neural Network (DCRNN; \citealt{Li2018DiffusionCRNN}), and Graph Attention Spatial-Temporal Network (GASTN; \citealt{He2020GraphAttention}), and Multi-Scale (MS) options such as MS-STGCN (~\citealt{Chen2021MultiScaleSTGCN}) and MS Graph Network (MSGNET; \citealt{Cai2024MsgNet}). Appendix~\ref{appendix:regional} explores additional baseline methods for the special case of downscaling from a regionally trained model to a local watershed.

\paragraph*{Performance comparison.}
Table~\ref{tab:table1} compares the predictive performance of MSGL with several baseline models across different watersheds using 1\% of fine-scale observations randomly selected from the training period. The multi-scale models (MS-STGCN, MSGNET, MSGL, and ASYNC-MSGL) perform substantially better in sparse scenarios than the single-scale models (RGrN, STGCN, GASTN, and DCRNN). Among the multi-scale models, our MSGL and ASYNC-MSGL consistently exhibit superior performance in all four watersheds, demonstrating the robustness and adaptability of our approach across datasets.

 We further evaluate the contributions of each component (CSL, CrSL, FSL, and MSO) through an ablation study. The combination of CSL and FSL ("MSGL (w/o CrSL)") improves performance over individual CSL and FSL but does not achieve the performance of the complete MSGL model. This indicates the importance of CrSL for capturing the relationship across scales. Similarly, the combination of CrSL and FSL ("MSGL (w/o CSL)") outperforms individual tasks but still falls short compared to the complete MSGL model. This highlights the critical role of CSL in capturing long-distance patterns and general water temperature dynamics  at the coarse scale, essential for enhancing the overall predictive accuracy. 
For the optimization process, the comparison between "MSGL" and "MSGL (w/o MSO)" indicates that MSO substantially enhances performance for the Lower West Branch Delaware, Neversink, and Rancocas watersheds, and provides a slight improvement for Lower Delaware. This suggests that MSO contributes to downscaling across different watersheds and highlights the overall benefit of incorporating MSO in the optimization process. 

As for MSGL and ASYNC-MSGL, adding the asynchronous strategy substantially improves performance across most watersheds and two data sparsity levels (Table~\ref{tab:table1} at 1\% of fine-scale labels and Figure~\ref{fig:spatialRMSE} at 0.1\%), with the Rancocas watershed at 1\% the only exception.

\subsection{Robustness under different  sparsity levels \textbf{(RQ2)}}
\label{subsection:experiment2}

To simulate label sparsity in these unusually well-observed study watersheds, fine-scale observations were randomly masked from all of each watershed's reach-date combinations (Appendix~\ref{appendix:details of data}).
In Figure~\ref{fig:sparseRMSE}, we compare robustness under data sparsity among the six spatiotemporal graph models introduced in RQ1-Baselines and our two new methods. Whereas MSGL and ASYNC-MSGL perform similarly to baseline methods when fine-scale labels are abundant (trained on 20\%-100\% of original observations), they outperform the baseline methods when labels are sparse (0.1\%-20\% of original observations). Specifically, as the fine-scale labels become sparser, the RMSE of MSGL and ASYNC-MSGL increases more slowly than for other methods. In extremely sparse environments, such as with only 0.1\% of observations (which is representative of real-world downscaling tasks), ASYNC-MSGL consistently and substantially outperforms other methods, demonstrating robustness to data sparsity.

\subsection{Comparison of cross-domain pretraining and Async-MSGL \textbf{(RQ3)}}

\label{subsection:experiment3}
Cross-domain pretraining is a common pre-training strategy aimed at improving the generalization ability of models. We utilize data from the Brandywine-Christina River Watershed, as shown in Figure~\ref{fig:studyarea}, for cross-domain pre-training. Results in Table~\ref{tab:RQ3} suggest that pretraining on $\mathcal{D}_{\text{mapping}}$ 
(ASYNC-MSGL) generally provides better performance across different watersheds, particularly in scenarios with very limited fine-scale data (0.1\%). Although cross-domain pretraining can substantially benefit downscaling on certain sparse datasets, these improvements are inconsistent; in many cases, it may lead to slight degradation or negligible change in performance. In contrast, our ASYNC-MSGL enhances model performance more consistently across watersheds and data sparsity scenarios.

The effectiveness of any stream temperature downscaling approach hinges on (1) the existence of correlations between coarse- and fine-scale information, such that values predicted at a coarse scale, when remapped to a finer scale, accurately reflect the general patterns of temperature variation, and (2) that the finer-resolution inputs and observations contain additional information about temperature variation not present in the coarse data. The asynchronous strategy further assumes that (3) coarse predictions made from a broader-extent model reflect additional learned information not present in the local data at the fine or coarse resolution. The success of ASYNC-MSGL in these experimental results, even when fine-scale observations are sparse, validates these three assumptions.

%% file: sections/06_discussion.tex
\section{Conclusion} \label{conclusion}

We introduce MSGL to address the challenges of fine-scale stream temperature prediction where data are sparse. Existing general multi-resolution approaches fail to consider the graph structure of water temperature data at different scales. Building on a hydrological understanding of stream graph structure, we propose a cross-scale learning task that incorporates positional information between nodes at different scales, enabling the model to capture physical connections across scales. We further enrich MSGL with an asynchronous strategy for pretraining on predictions from a coarse-resolution regional model, \( \mathcal{D}_{\text{mapping}} \), to leverage relationships learned on regional datasets for local prediction needs. Extensive experiments in the Delaware River Basin demonstrate the superior predictive accuracy of MSGL and ASYNC-MSGL over six baseline single-scale and multi-scale graph models, especially in scenarios of data sparsity. The methods presented here are readily transferable to other cases where data are irregularly structured and sparsely observed at the target resolution, such as prediction of other water quality variables or modeling of meteorological dynamics in complex terrain.



%% file: sections/07_acknowledgements.tex
\section*{Acknowledgements}
YF, RY, and XJ were supported by the National Science Foundation (NSF) under grants 2239175, 2316305, 2147195, and 2425845, the USGS awards  G21AC10564 and G22AC00266, and the NASA grant 80NSSC24K1061. 
YX was supported by the NSF under grants 2126474, 2147195, 2425844, 2430978, the Google’s AI for Social Good Impact Scholars program. This research was also supported in part by the University of Pittsburgh Center for Research Computing through the resources provided. 
JRB was supported by the U.S. Department of Energy, Office of Science, Office of Biological and Environmental Research, Environmental System Science Data Management Program, as part of the ExaSheds project, under Award Number 89243021SSC000068. APA was supported by the U.S. Geological Survey, Water Availability and Use Science Program (USGS WAUSP). We thank Lauren Koenig for sharing prepared data and insights into the stream networks at the fine resolution studied here. We thank Jeremy Diaz, Scott Hamshaw, and several anonymous reviewers for their thoughtful reviews of the manuscript. 
Any use of trade, firm, or product names is for descriptive purposes only and does not imply endorsement by the U.S. Government. Model inputs, code, and outputs are provided in Barclay et al.~\citeyear{Barclay2024MSGLData} (https://doi.org/10.5066/P1UP5DXN).

%% file: sections/09_appendix.tex
\appendix
\setcounter{secnumdepth}{2} 

\section{Details of Multi-Scale Graph Learning}
\subsection{Multi-Scale Optimization (MSO) algorithm}
The MSO algorithm is a extension of the Pareto Multi-Objective Optimization Algorithm and Multiple-Gradient Descent Algorithm (MGDA)~\cite{Deb2005RobustPareto, Desideri2012MGDA, Ozan2018MultiTaskLearning, Mahapatra2020ParetoMTL, Suzuki2020MultiObjectiveBO} and Robust Multi-Objective Optimization Algorithm~\cite{Roberts2018RobustHybridPower, Zhou2018MultiObjectiveRobust, Daulton2022RobustMOBO}.

\begin{algorithm}[H]
\label{appendix:MGDA}
\caption{MGDA-Based Multi-Scale Optimization (MSO)}
\label{alg:generalized_solver}
\begin{algorithmic}[1]
\Require Shared parameters $\Theta$, task-specific parameters $\Phi_{c}$, $\Phi_{cr}$, and $\Phi_{f}$.
\State Initialize shared parameters $\Theta$ and task-specific parameters $\{\Phi_{c}, \Phi_{cr}, \Phi_{f}\}$
\For{$t \in \{c, cr, f\}$}
    \State $\Phi_{\tau} = \Phi_{\tau} - \eta \nabla_{\Phi_{\tau}} \lL_{{\tau}}(\Theta, \Phi_{{\tau}})$  \Comment{Gradient descent for specific-scale learning parameters}
\EndFor
\State
\State $\vec{\nabla} \gets [\nabla_{\Theta} \lL_{c}, \nabla_{\Theta} \lL_{cr}, \nabla_{\Theta} \lL_{f}]$
\State Compute weights $\ensuremath{\vec{\alpha} = \left( \alpha_{\tau} \right)_{\tau \in \{\text{c}, \text{cr}, \text{f}\}}}$ using \textproc{MGDA}{$(\vec{\nabla})$} 

\State $\sum_{\tau \in \{c, cr, f\}} \alpha_{\tau} \nabla_{\Theta} \lL_{\tau}$ \Comment{Compute the weighted sum of gradients for shared parameters}

\State $\Theta = \Theta - \eta \sum_{\tau \in \{c, cr, f\}} \alpha_{\tau} \nabla_{\Theta} \lL_{\tau}$ \Comment{Update shared parameters $\Theta$ based on weighted gradients}

\State
\State

\Procedure{MGDA}{$\vec{\nabla}$}
    \State Initialize weights $\bm{\alpha} = \left(\frac{1}{3}, \frac{1}{3}, \frac{1}{3}\right)$ \Comment{Three scale-learning tasks}
    \State Precompute pairwise scalar products $\MM_{i,j} = \vec{\nabla}_i^\intercal \vec{\nabla}_j$ for all $i,j \in \{c, cr, f\}$
    \Repeat
        \State $\hat{t} = \argmin_r \sum_t \alpha^t \MM_{rt}$ \Comment{Find the task with minimal contribution}
        \State $\btheta = \MM \bm{\alpha}$
        \State $\bar{\btheta} = \MM \bm{e}_{\hat{t}}$
        \Comment{\(\bm{e}_{\hat{t}}\) is the unit vector in the \(\hat{t}\) direction}
        \If{$\btheta^\intercal \bar{\btheta} \geq \btheta^\intercal \btheta$}
            \State $\hat{\gamma} = 1$ \Comment{Set $\gamma$ to 1 if contribution from $\bar{\btheta}$ is maximal}
        \ElsIf{$\btheta^\intercal \bar{\btheta} \geq \bar{\btheta}^\intercal \bar{\btheta}$}
            \State $\hat{\gamma} = 0$ \Comment{Set $\gamma$ to 0 if contribution from $\btheta$ is maximal}
        \Else
            \State $\hat{\gamma} = \frac{(\bar{\btheta} - \btheta)^\intercal \bar{\btheta}}{\|\btheta - \bar{\btheta}\|_2^2}$ \Comment{Calculate optimal $\gamma$ based on both contributions}
        \EndIf
        \State $\bm{\alpha} = (1- \hat{\gamma})\bm{\alpha} + \hat{\gamma} \bm{e}_{\hat{t}}$ 
        \Comment{Update weights towards \(\hat{t}\) direction}
    \Until{$\hat{\gamma} \sim 0$ \textbf{or} iteration limit}
    \State \textbf{return} $\bm{\alpha}$ \Comment{Return weights $\alpha$ for each scale learning}
\EndProcedure
\end{algorithmic}
\end{algorithm}

\vfill
\subsection{Cross-scale distance matrix}

Let \(A\) and \(B\) be nodes at a coarse scale, and denote by \(\mathcal{N}_{\epsilon}(B)\) the \(\epsilon\)-neighborhood of \(B\) at a finer scale, where \(\epsilon\) is a small positive value representing the radius of the neighborhood. If the relative position of \(A\) to \(B\) at the coarse scale is given by a distance function \(d_{\text{coarse}}(A, B)\), then the relative positions at the finer scale can be approximated by the set of distances \(\{d_{\text{cross}}(A, B')\}_{B' \in \mathcal{N}_{\epsilon}(B)}\), where \(B'\) represents nodes within the neighborhood \(\mathcal{N}_{\epsilon}(B)\). These approximations are encapsulated within the cross-scale distance matrix \(\mathbf{D}_{\text{cross}}\), which plays a crucial role in our MSGL approach.

\textbf{Lemma}: For nodes \(A\) and \(B\) in the coarse-scale graph, and the \(\epsilon\)-neighborhood \(\mathcal{N}_{\epsilon}(B)\) in the fine-scale graph, if \(\epsilon\) is sufficiently small, there exists a function \(K(\epsilon) > 0\) that decreases as \(\epsilon\) decreases, such that for all \(B' \in \mathcal{N}_{\epsilon}(B)\), we have:
\[
\left| d_{\text{coarse}}(A, B) - d_{\text{cross}}(A, B') \right| \leq K(\epsilon).
\]
Here, \(K(\epsilon)\) is a function that depends on \(\epsilon\) and the structure and metric of the graphs.

\textbf{Proposition}: As \(\epsilon \to 0\), the difference between the maximum and minimum elements of the cross-scale distance matrix \(\mathbf{D}_{\text{cross}}(A, B)\) approaches zero. Formally:
\[
\lim_{\epsilon \to 0} \left( \max_{B' \in \mathcal{N}_{\epsilon}(B)} d_{\text{cross}}(A, B') - \min_{B' \in \mathcal{N}_{\epsilon}(B)} d_{\text{cross}}(A, B') \right) = 0.
\]

\textbf{Proof}: According to the Lemma, when \(\epsilon\) is sufficiently small, for all \(B' \in \mathcal{N}_{\epsilon}(B)\), we have:
\[
\left| d_{\text{coarse}}(A, B) - d_{\text{cross}}(A, B') \right| \leq K(\epsilon).
\]
Here, the function \(K(\epsilon)\) decreases as \(\epsilon\) decreases. Therefore, for all \(B' \in \mathcal{N}_{\epsilon}(B)\), the cross-scale distance \(d_{\text{cross}}(A, B')\) fluctuates around the coarse-scale distance \(d_{\text{coarse}}(A, B)\), with the fluctuation bounded by \(K(\epsilon)\).

By the Lemma, we have:
\[
d_{\text{coarse}}(A, B) - K(\epsilon) \leq d_{\text{cross}}(A, B') \leq d_{\text{coarse}}(A, B) + K(\epsilon).
\]
Thus:
\[
\min_{B' \in \mathcal{N}_{\epsilon}(B)} d_{\text{cross}}(A, B') \geq d_{\text{coarse}}(A, B) - K(\epsilon),
\]
\[
\max_{B' \in \mathcal{N}_{\epsilon}(B)} d_{\text{cross}}(A, B') \leq d_{\text{coarse}}(A, B) + K(\epsilon).
\]
Hence, the difference between the maximum and minimum values is:
\[
\max_{B' \in \mathcal{N}_{\epsilon}(B)} d_{\text{cross}}(A, B') - \min_{B' \in \mathcal{N}_{\epsilon}(B)} d_{\text{cross}}(A, B') \leq 2K(\epsilon).
\]

As \(\epsilon \to 0\), the range of \(B'\) within the \(\epsilon\)-neighborhood shrinks to zero, meaning that all values of \(d_{\text{cross}}(A, B')\) converge to the same value, which is \(d_{\text{coarse}}(A, B)\). Because \(K(\epsilon)\) decreases as \(\epsilon\) decreases, the difference between the maximum and minimum values tends to zero:
\[
\lim_{\epsilon \to 0} \left( \max_{B' \in \mathcal{N}_{\epsilon}(B)} d_{\text{cross}}(A, B') - \min_{B' \in \mathcal{N}_{\epsilon}(B)} d_{\text{cross}}(A, B') \right) = 0.
\]
This completes the proof of the Proposition.

\clearpage

\section{Details of Data}\label{appendix:details of data}

All model inputs, code, and results are provided in Barclay et al.~\citeyear{Barclay2024MSGLData}. An overview of the datasets and our approach to data preparation is given in this appendix.

\subsection{Data characteristics}\label{appendix:data characteristics}
\paragraph*{Coarse-scale and Fine-scale.}
\label{Coarse-scale and Fine-scale.}
Table~\ref{table:watersheds} shows five watersheds selected from the Delaware River Basin for our experiments. These variables were included as predictors of water temperature: daily precipitation, daily mean air temperature, mean reach elevation, incoming short-wave radiation, streambed slope, mean reach width, potential evapotranspiration. Descriptions of these variables are given in~\cite{Terry2022StreamTemperatureData, Oliver2022WaterQualityData}.

\newcolumntype{L}[1]{>{\raggedright\arraybackslash}p{#1}}
\newcolumntype{C}[1]{>{\centering\arraybackslash}X}
\newcolumntype{R}[1]{>{\raggedleft\arraybackslash}X}
\begin{table}[ht] 
    \centering
    \caption{Details of selected watersheds in the Delaware River Basin, showing the number of observed stream reaches and the average observations per reach, differentiated by coarse and fine classification.}
    \label{table:watersheds}
    \begin{tabularx}{\linewidth}{L{2.3cm}C{1.2cm}C{1.2cm}C{1.2cm}C{1.2cm}C{1.2cm}}
        \toprule
        Dataset & \rotatebox{270}{Lower Delaware} &  \rotatebox{270}{LW Branch Delaware} & \rotatebox{270}{Neversink} & \rotatebox{270}{Rancocas} & \rotatebox{270}{Brandywine-Christina} \\
        \midrule
        \footnotesize Number of Coarse Reaches & 9 & 5 & 11 & 11 & 22 \\
        \cmidrule(lr){1-6}
        \footnotesize Number of Fine Reaches & 18 & 11 & 23 & 39 & 46 \\
        \cmidrule(lr){1-6}
        \footnotesize Labels per Coarse Reach & 2335 & 7893 & 2524 & 1159 & 2848 \\
        \cmidrule(lr){1-6}
        \footnotesize Labels per Fine Reach & 1110 & 4489 & 1369 & 343 & 1372 \\
        \bottomrule
    \end{tabularx}
\end{table}

\subsection{Data partitioning}
\label{appendix:data partition}
\begin{figure}[H]
    \centering
    \includegraphics[width=\linewidth]{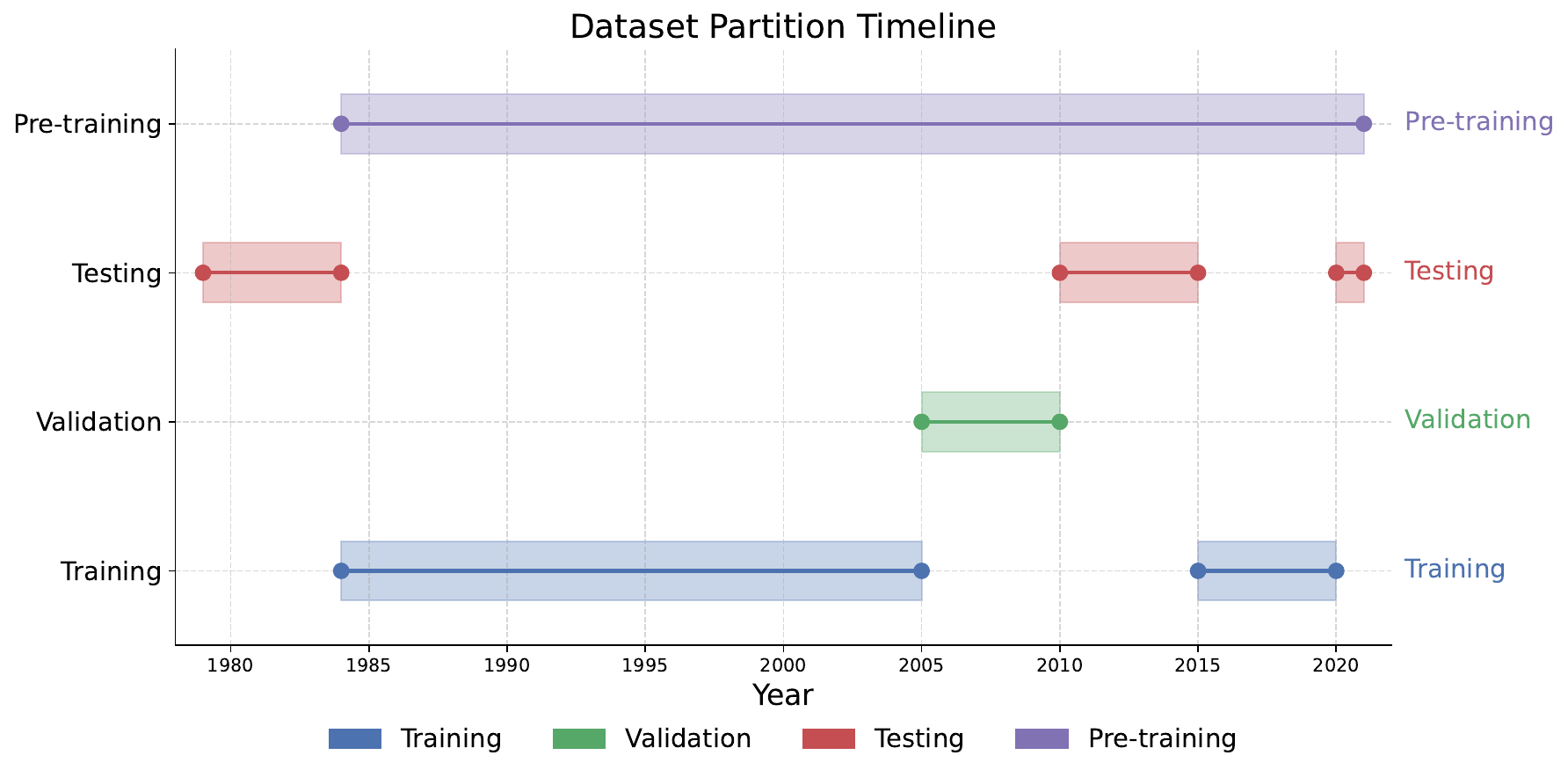}
    \caption{Timeline of data partitions in Lower Delaware, Neversink, Lower West Branch Delaware, and Rancocas Watersheds.}
    \label{fig:datapartitions}
\end{figure}

The dataset spans several decades and is divided into distinct sets for training, validation, testing, and pre-training to support the development and evaluation of predictive models~\ref{fig:datapartitions}.

The data collection and partitioning were conducted across five watersheds: Brandywine, Lower Delaware, Neversink, Lower West Branch Delaware, and Rancocas. Each of these watersheds provided a unique set of environmental conditions and hydrological data, which were crucial for ensuring that the models developed are robust and capable of generalizing across different geographical and climatic conditions.

For the pre-training phase of our models, a separate dataset was utilized, distinct from the training, validation, and testing sets described earlier. This dataset was specifically collected from the Christina River Watershed, which is not one of the five watersheds used in the main dataset. The Christina River Watershed provides a diverse and complementary environmental setting, which is ideal for cross-region pre-training.

By leveraging data from a different geographical region, our approach ensures that the predictive models are not only tailored to specific local conditions but also capable of generalizing across broader regional variations. This cross-region pre-training strategy helps in developing more robust models that can perform well under various hydrological and environmental conditions not covered by the initial training datasets.

This segmentation ensured comprehensive coverage and utility across different temporal segments, providing a robust basis for model training and evaluation. For further details on the experiments conducted using this dataset, refer to the discussion of RQ3.

\subsection{Simulating data sparsity}
\label{appendix:datasparsity}
This section provides supplementary details about the dataset used in experiments RQ1, RQ2, RQ3, and Appendix~\ref{appendix:regional}.

To simulate real-world scenarios of data sparsity, the fine-scale labels within the train and validation datasets were subjected to a masking process. This involved retaining a controlled percentage of observational labels, such as 0.1\%, 1\%, and 100\%, mimicking conditions ranging from severe data scarcity to full data availability. In our Delaware datasets, full data availability means that all available observations were used, though this was typically only a small fraction of the number of reaches times the number of days in the 1979-2021 period (roughly 1-6\%). This approach not only tests the robustness of the predictive models under different data sparsity levels but also mirrors challenges encountered in practical environmental monitoring and data collection efforts. Simulating data sparsity gives us insights into how these models might perform under less-than-ideal, yet common, conditions.

\begin{figure}[H]
    \centering
    \includegraphics[width=\linewidth]{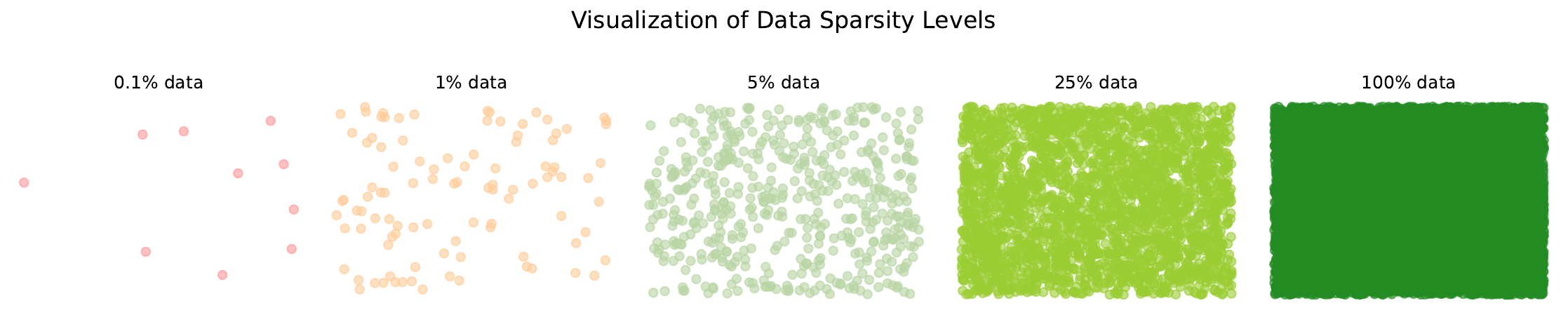}
    \caption{Sparsity levels visualized for training and validation datasets at 0.1\%, 1\%, 5\%, 25\%, and 100\% data retention. This chart demonstrates the range of conditions from minimal to full data availability.}
    \label{fig:pretrainingdataset}
\end{figure}

By adjusting the reduction percentage parameter, users can simulate realistic scenarios where data might be less available than in the well-observed watersheds of our study.

To achieve this simulation, we employed a systematic data reduction method. The process begins by reading the dataset and converting it into a tabular format, ordered by date. For specific training and validation periods, we identified the relevant data segments. Within each period, we set a random seed (for reproducibility) and then randomly selected a specified proportion of data points and marked them as missing to simulate real-world data gaps. This random selection ensured unbiased reduction and reproducibility. The masked dataset was then converted back to its original format for further analysis. This method supports data reduction for datasets with multiple variables and can be applied selectively to specific segments. The final dataset, reflecting the sparse data, was used for model training. Unmasked data were used for model evaluation.

\section{Experimental Setting}
\subsection{Model hyperparameters}
\paragraph{MSGL:}
{Input Dimension}: The model uses an input dimension of 16.\\
{Hidden Dimension}: The hidden state size is set to 64. \\
{Recurrent Dropout)}: A dropout rate of 0.2 is applied to the recurrent layers. \\
{Dropout}: A dropout rate of 0.5 is applied to the input layers. \\
{Seed}: For reproducibility, the model is initialized with random seeds of 1, 2, or 3. These seeds were selected arbitrarily and were not modified after experimentation began. \\
{Number of Attention Heads)}: The model uses 4 attention heads in the multihead attention layers. \\
{Task1 Module Dropout}: A specific dropout rate of 0.1 is applied within the Task1Module for additional regularization. \\

\subsection{Training settings}
{Early Stopping}: Training is stopped if the performance does not improve for 30 consecutive epochs. \\
{Pre-training Epochs}: The model undergoes 60 epochs of pre-training. \\
{Fine-tuning Epochs}: The model undergoes 60 epochs of fine-tuning. \\
{Pre-training Learning Rate}: The initial learning rate for pre-training is set to 0.005, with scheduled 70\% reductions in the learning rate at 40 and 50 epochs. \\
{Fine-tuning Learning Rate}: TThe initial learning rate for fine-tuning is set to 0.005, with scheduled 70\% reductions in the learning rate at 40 and 50 epochs. \\
{Random Seeds for Label Mask}: The fine-scale dataset's label is masked with three different random seeds: 42, 61, and 71. These seeds were selected arbitrarily and were not modified after experimentation began.

\subsection{Computational environments}
The experiments were conducted using the following computational resources:

The hardware used includes NVIDIA RTX 4080 GPUs with 16 GB memory, 13th Gen Intel(R) Core(TM) i9-13900KF running at 3.0GHz, 32 GB of DDR4 RAM, and 2 TB SSD storage.

The software environment consisted of Windows 11 Enterprise Edition as the operating system, with PyTorch 1.8.2 serving as the deep learning framework. Additionally, CUDA version 11.1.1 and cuDNN version 8.0 were utilized to optimize GPU performance. Other essential libraries included NumPy 1.21.5 and SciPy 1.7.3.

\section{Baselines for Downscaling from Regional Models}
\label{appendix:regional}

A special case of the broader downscaling challenge is to make use of a regionally trained model to predict in a local watershed. ASYNC-MSGL offers one approach to this special case; while ASYNC-MSGL is also well suited to downscaling challenges where only local data (coarse and fine scale) are available, in Table~\ref{table:regional-down} we investigate the performance of ASYNC-MSGL against other baseline approaches specifically tailored to the special case of downscaling from a regional model.

The tailored baseline approaches we consider are:

\paragraph*{Repredict} A method that applies the trained regional coarse-scale model (i.e., architecture, weights, and scaling factors for input variables), using as new input the rescaled fine-scale input data from the watershed of interest to produce fine-resolution predictions.

\paragraph*{Refine} A fine-scale learning approach for the watershed of interest that incorporates remapped coarse predictions from the regional coarse-scale model as an additional input variable. The coarse predictions are first remapped to the fine-scale reaches using a simple lookup table of which coarse-scale reach is spatially coincident with each fine-scale reach. The downscaling model is then pretrained on the regional coarse-scale model outputs and subsequently fine-tuned on the fine-scale observations.

\paragraph*{Residuals} After remapping coarse predictions from the regional coarse-scale learning onto the fine-scale reaches, as described under Refine, residuals between the fine-scale (remapped) predictions and fine-scale observations are calculated. A fine-scale learning approach for the watershed of interest is then trained on these residuals, and the predicted residuals are added to the remapped predictions to predict water temperatures.

\paragraph*{} Table~\ref{table:regional-down} compares these three regional-to-local downscaling methods to three methods described in the main manuscript: \textbf{D$_{mapping}$}, the predictions from the regional coarse-scale RGrN remapped to fine-scale reaches; \textbf{RGrN}, a fine-scale-only application of the RGrN method; and \textbf{ASYNC-MSGL}, which uses the MSGL architecture, pretrains on \textbf{D$_{mapping}$}, and fine-tunes on fine-scale data.

\begin{table} [ht]
    \centering
    \caption{Methods comparison at 1\% of fine-scale observations used for training (and 1\% of coarse-scale observations used for training for Refine and Residuals) for $D_{mapping}$, RGrN, ASYNC-MSGL, and the three baseline regional-downscaling methods described in this appendix.}
    \label{table:regional-down}
    \begin{tabularx}{\linewidth}{L{1.5cm}C{2cm}C{2cm}C{2cm}C{2cm}}
        \toprule
        Method & \rotatebox{270}{Lower Delaware} &  \rotatebox{270}{LW Branch Delaware} & \rotatebox{270}{Neversink} & \rotatebox{270}{Rancocas} \\
        \cmidrule{1-5}
        Repredict & 9.035 $\pm$ 0.000 & 4.597 $\pm$ 0.000 & 4.531 $\pm$ 0.000 & 2.253 $\pm$ 0.000\\
        \cmidrule{1-5}
        Refine & 1.239 $\pm$ 0.090 & 2.241 $\pm$ 0.140 & 1.852 $\pm$ 0.179 & 1.786 $\pm$ 0.074\\
        \cmidrule{1-5}
        Residuals & 1.206 $\pm$ 0.069 & 2.180 $\pm$ 0.057 & 1.772 $\pm$ 0.136 & 1.599 $\pm$ 0.119\\
        \cmidrule{1-5}
        D$_{mapping}$ & 1.144 $\pm$ 0.000 & 2.104 $\pm$ 0.000 & 1.988 $\pm$ 0.000 & 1.417 $\pm$ 0.000\\
        \cmidrule{1-5}
        RGrN & 1.732 $\pm$ 0.131 & 2.648 $\pm$ 0.154 & 1.935 $\pm$ 0.090 & 2.198 $\pm$ 0.246\\
        \cmidrule{1-5}
        ASYNC-MSGL & 1.360 $\pm$ 0.116 & 2.179 $\pm$ 0.129 & 1.617 $\pm$ 0.134 & 1.483 $\pm$ 0.053\\
        \bottomrule
    \end{tabularx}
\end{table}

\clearpage

%% file: aaai25.bbl
\begin{thebibliography}{60}
\providecommand{\natexlab}[1]{#1}

\bibitem[{A.~Mullen and Bratt(2018)}]{a_mullen_usaboundaries_2018}
A.~Mullen, L.; and Bratt, J. 2018.
\newblock {USAboundaries}: {Historical} and {Contemporary} {Boundaries} of the {United} {States} of {America}.
\newblock \emph{Journal of Open Source Software}, 3(23): 314.

\bibitem[{Abatzoglou(2013)}]{gridmet2013}
Abatzoglou, J.~T. 2013.
\newblock Development of gridded surface meteorological data for ecological applications and modelling.
\newblock \emph{International Journal of Climatology}, 33(1): 121--131.

\bibitem[{Barclay et~al.(2024)Barclay, Fan, Koenig, Yu, Sun, Xie, Jia, and Appling}]{Barclay2024MSGLData}
Barclay, J.~R.; Fan, Y.; Koenig, L.; Yu, R.; Sun, Y.; Xie, Y.; Jia, X.; and Appling, A.~P. 2024.
\newblock Downscaling and multi-scale modeling of stream temperature in five watersheds of the Delaware River Basin, 1979-2021.
\newblock In \emph{U.S. Geological Survey data release}.

\bibitem[{Barclay et~al.(2023)}]{Barclay2023ProcessGuidedDL}
Barclay, J.~R.; et~al. 2023.
\newblock Train, Inform, Borrow, or Combine? Approaches to Process‐Guided Deep Learning for Groundwater‐Influenced Stream Temperature Prediction.
\newblock \emph{Water Resources Research}, 59(12): e2023WR035327.

\bibitem[{Bock et~al.(2020)Bock, Santiago, Wieczorek, Foks, and Lombard}]{bock_geospatial_2020}
Bock, A.~R.; Santiago, M.; Wieczorek, M.~E.; Foks, S.~S.; and Lombard, M.~A. 2020.
\newblock Geospatial {Fabric} for {National} {Hydrologic} {Modeling}, version 1.1.

\bibitem[{Brett(1971)}]{brett1971energetic}
Brett, J.~R. 1971.
\newblock Energetic responses of salmon to temperature. A study of some thermal relations in the physiology and freshwater ecology of sockeye salmon (Oncorhynchus nerkd).
\newblock \emph{American Zoologist}.

\bibitem[{Cai(2024)}]{Cai2024MsgNet}
Cai, W. e.~a. 2024.
\newblock MSGNet: Learning Multi-Scale Inter-series Correlations for Multivariate Time Series Forecasting.
\newblock In \emph{Proceedings of the AAAI Conference on Artificial Intelligence}.

\bibitem[{Chau, Bouabid, and Sejdinovic(2021)}]{Chau2021sta}
Chau, S.~L.; Bouabid, S.; and Sejdinovic, D. 2021.
\newblock Deconditional downscaling with gaussian processes.
\newblock In \emph{Advances in Neural Information Processing Systems}, volume~34, 17813--17825.

\bibitem[{Chaudhuri and Robertson(2020)}]{Chaudhuri2020sta}
Chaudhuri, C.; and Robertson, C. 2020.
\newblock CliGAN: A Structurally Sensitive Convolutional Neural Network Model for Statistical Downscaling of Precipitation from Multi-Model Ensembles.
\newblock \emph{Water}, 12(12): 3353.

\bibitem[{Chen, Fan, and Panda(2021)}]{Chen2021CrossViT}
Chen, C.-F.~R.; Fan, Q.; and Panda, R. 2021.
\newblock CrossViT: Cross-attention multi-scale vision transformer for image classification.
\newblock In \emph{Proceedings of the IEEE/CVF International Conference on Computer Vision}.

\bibitem[{Chen et~al.(2021)Chen, Appling, Oliver, Corson-Dosch, Read, Sadler, Zwart, and Jia}]{chen2021heterogeneous}
Chen, S.; Appling, A.; Oliver, S.; Corson-Dosch, H.; Read, J.; Sadler, J.; Zwart, J.; and Jia, X. 2021.
\newblock Heterogeneous stream-reservoir graph networks with data assimilation.
\newblock In \emph{2021 IEEE International Conference on Data Mining (ICDM)}, 1024--1029. IEEE.

\bibitem[{Chen et~al.(2023)Chen, Kalanat, Xie, Li, Zwart, Sadler, Appling, Oliver, Read, and Jia}]{chen2023physics}
Chen, S.; Kalanat, N.; Xie, Y.; Li, S.; Zwart, J.~A.; Sadler, J.~M.; Appling, A.~P.; Oliver, S.~K.; Read, J.~S.; and Jia, X. 2023.
\newblock Physics-guided machine learning from simulated data with different physical parameters.
\newblock \emph{Knowledge and Information Systems}, 65(8): 3223--3250.

\bibitem[{Chen, Zwart, and Jia(2022)}]{chen2022physics}
Chen, S.; Zwart, J.~A.; and Jia, X. 2022.
\newblock Physics-guided graph meta learning for predicting water temperature and streamflow in stream networks.
\newblock In \emph{Proceedings of the 28th ACM SIGKDD Conference on Knowledge Discovery and Data Mining}, 2752--2761.

\bibitem[{Chen(2021)}]{Chen2021MultiScaleSTGCN}
Chen, Z. e.~a. 2021.
\newblock Multi-Scale Spatial Temporal Graph Convolutional Network for Skeleton-Based Action Recognition.
\newblock In \emph{Proceedings of the AAAI Conference on Artificial Intelligence}.

\bibitem[{Daulton et~al.(2022)}]{Daulton2022RobustMOBO}
Daulton, S.; et~al. 2022.
\newblock Robust multi-objective bayesian optimization under input noise.
\newblock In \emph{International Conference on Machine Learning}. PMLR.

\bibitem[{Deb and Gupta(2005)}]{Deb2005RobustPareto}
Deb, K.; and Gupta, H. 2005.
\newblock Searching for robust pareto-optimal solutions in multi-objective optimization.
\newblock In \emph{Evolutionary Multi-Criterion Optimization}, 150--164. Springer Berlin Heidelberg.

\bibitem[{Di~Giovanni et~al.(2023)Di~Giovanni, Giusti, Barbero, Luise, Lio, and Bronstein}]{di2023over}
Di~Giovanni, F.; Giusti, L.; Barbero, F.; Luise, G.; Lio, P.; and Bronstein, M.~M. 2023.
\newblock On over-squashing in message passing neural networks: The impact of width, depth, and topology.
\newblock In \emph{International Conference on Machine Learning}, 7865--7885. PMLR.

\bibitem[{Désidéri(2012)}]{Desideri2012MGDA}
Désidéri, J.-A. 2012.
\newblock Multiple-gradient descent algorithm (MGDA) for multiobjective optimization.
\newblock \emph{Comptes Rendus Mathematique}, 313--318.

\bibitem[{Fotakis(2021)}]{Fotakis2021CoarseLabels}
Fotakis, D. e.~a. 2021.
\newblock Efficient algorithms for learning from coarse labels.
\newblock In \emph{Conference on Learning Theory}. PMLR.

\bibitem[{Gu et~al.(2022)}]{Gu2022HighResVisionTransformer}
Gu, J.; et~al. 2022.
\newblock Multi-scale high-resolution vision transformer for semantic segmentation.
\newblock In \emph{Proceedings of the IEEE/CVF Conference on Computer Vision and Pattern Recognition}.

\bibitem[{Hamelijnck et~al.(2019)Hamelijnck, Damoulas, Wang, and Girolami}]{Hamelijnck2019aggregate}
Hamelijnck, O.; Damoulas, T.; Wang, K.; and Girolami, M.~A. 2019.
\newblock Multi-resolution multi-task Gaussian processes.
\newblock In \emph{Advances in Neural Information Processing Systems}.

\bibitem[{Harmel et~al.(2023)Harmel, Preisendanz, King, Busch, Birgand, and Sahoo}]{Harmel2023WQCosts}
Harmel, R.~D.; Preisendanz, H.~E.; King, K.~W.; Busch, D.; Birgand, F.; and Sahoo, D. 2023.
\newblock A Review of Data Quality and Cost Considerations for Water Quality Monitoring at the Field Scale and in Small Watersheds.
\newblock \emph{Water}, 15(17).

\bibitem[{He et~al.(2024)He, Xie, Sun, Zwart, Yang, Jin, Wang, Karimi, and Jia}]{he2024fair}
He, E.; Xie, Y.; Sun, A.; Zwart, J.; Yang, J.; Jin, Z.; Wang, Y.; Karimi, H.; and Jia, X. 2024.
\newblock Fair Graph Learning Using Constraint-Aware Priority Adjustment and Graph Masking in River Networks.
\newblock In \emph{Proceedings of the AAAI Conference on Artificial Intelligence}, volume~38, 22087--22095.

\bibitem[{He(2020)}]{He2020GraphAttention}
He, K. e.~a. 2020.
\newblock Graph attention spatial-temporal network with collaborative global-local learning for citywide mobile traffic prediction.
\newblock \emph{IEEE Transactions on Mobile Computing}, 21(4): 1244--1256.

\bibitem[{He, Li, and Chen(2017)}]{He2017Multiscale3DCNN}
He, M.; Li, B.; and Chen, H. 2017.
\newblock Multi-scale 3D deep convolutional neural network for hyperspectral image classification.
\newblock In \emph{2017 IEEE International Conference on Image Processing (ICIP)}. IEEE.

\bibitem[{Hochreiter and Schmidhuber(1997)}]{hochreiter_long_1997}
Hochreiter, S.; and Schmidhuber, J. 1997.
\newblock Long {Short}-{Term} {Memory}.
\newblock \emph{Neural Computation}, 9(8): 1735--1780.

\bibitem[{Ji, Zhong, and Ma(2021)}]{Ji2021SuperResolution}
Ji, J.; Zhong, B.; and Ma, K.-K. 2021.
\newblock Single image super-resolution using asynchronous multi-scale network.
\newblock \emph{IEEE Signal Processing Letters}, 28: 1823--1827.

\bibitem[{Jia et~al.(2023)Jia, Chen, Zheng, Xie, Jiang, and Kalanat}]{jia2023physics}
Jia, X.; Chen, S.; Zheng, C.; Xie, Y.; Jiang, Z.; and Kalanat, N. 2023.
\newblock Physics-guided graph diffusion network for combining heterogeneous simulated data: An application in predicting stream water temperature.
\newblock In \emph{Proceedings of the 2023 SIAM International Conference on Data Mining (SDM)}, 361--369. SIAM.

\bibitem[{Jia et~al.(2021{\natexlab{a}})Jia, Xie, Li, Chen, Zwart, Sadler, Appling, Oliver, and Read}]{jia2021physics_simlr}
Jia, X.; Xie, Y.; Li, S.; Chen, S.; Zwart, J.; Sadler, J.; Appling, A.; Oliver, S.; and Read, J. 2021{\natexlab{a}}.
\newblock Physics-guided machine learning from simulation data: An application in modeling lake and river systems.
\newblock In \emph{2021 IEEE International Conference on Data Mining (ICDM)}, 270--279. IEEE.

\bibitem[{Jia et~al.(2021{\natexlab{b}})}]{Jia2021PhysicsGuidedRecurrent}
Jia, X.; et~al. 2021{\natexlab{b}}.
\newblock Physics-guided recurrent graph model for predicting flow and temperature in river networks.
\newblock In \emph{Proceedings of the 2021 SIAM International Conference on Data Mining (SDM)}. Society for Industrial and Applied Mathematics.

\bibitem[{Jiang et~al.(2020)}]{Jiang2020Deraining}
Jiang, K.; et~al. 2020.
\newblock Multi-scale progressive fusion network for single image deraining.
\newblock In \emph{Proceedings of the IEEE/CVF Conference on Computer Vision and Pattern Recognition}.

\bibitem[{Li et~al.(2018)Li, Yu, Shahabi, and Liu}]{Li2018DiffusionCRNN}
Li, Y.; Yu, R.; Shahabi, C.; and Liu, Y. 2018.
\newblock Diffusion convolutional recurrent neural network: Data-driven traffic forecasting.
\newblock In \emph{6th International Conference on Learning Representations (ICLR 2018)}.

\bibitem[{Li et~al.(2024)Li, Liu, Zhang, and Fu}]{li2024real}
Li, Z.; Liu, H.; Zhang, C.; and Fu, G. 2024.
\newblock Real-time water quality prediction in water distribution networks using graph neural networks with sparse monitoring data.
\newblock \emph{Water Research}, 250: 121018.

\bibitem[{Liu et~al.(2021)}]{Liu2021DeepFeatureEnsemble}
Liu, J.; et~al. 2021.
\newblock Learning a deep multi-scale feature ensemble and an edge-attention guidance for image fusion.
\newblock \emph{IEEE Transactions on Circuits and Systems for Video Technology}, 32(1): 105--119.

\bibitem[{Mahapatra and Rajan(2020)}]{Mahapatra2020ParetoMTL}
Mahapatra, D.; and Rajan, V. 2020.
\newblock Multi-task learning with user preferences: Gradient descent with controlled ascent in Pareto optimization.
\newblock In \emph{International Conference on Machine Learning}. PMLR.

\bibitem[{Moshe et~al.(2020)Moshe, Metzger, Elidan, Kratzert, Nevo, and El-Yaniv}]{moshe2020hydronets}
Moshe, Z.; Metzger, A.; Elidan, G.; Kratzert, F.; Nevo, S.; and El-Yaniv, R. 2020.
\newblock Hydronets: Leveraging river structure for hydrologic modeling.
\newblock \emph{arXiv preprint arXiv:2007.00595}.

\bibitem[{Oliver et~al.(2022)Oliver, Sleckman, Appling, Corson-Dosch, Zwart, Thompson, Koenig, White, Watkins, Platt, Padilla, and Sadler}]{Oliver2022WaterQualityData}
Oliver, S.; Sleckman, M.; Appling, A.; Corson-Dosch, H.; Zwart, J.; Thompson, T.; Koenig, L.; White, E.; Watkins, D.; Platt, L.; Padilla, J.; and Sadler, J. 2022.
\newblock Data to support water quality modeling efforts in the Delaware River Basin.

\bibitem[{Qin, Huang, and Wen(2020)}]{Qin2020SuperResolution}
Qin, J.; Huang, Y.; and Wen, W. 2020.
\newblock Multi-scale feature fusion residual network for single image super-resolution.
\newblock \emph{Neurocomputing}, 379: 334--342.

\bibitem[{Ravindranath et~al.(2016)}]{ravindranath2016environmental}
Ravindranath, A.; et~al. 2016.
\newblock An environmental perspective on the water management policies of the Upper Delaware River Basin.
\newblock \emph{Water Policy}.

\bibitem[{Regan et~al.(2018)Regan, Markstrom, Hay, Viger, Norton, Driscoll, and LaFontaine}]{Regan2018NationalHydrologicModel}
Regan, R.~S.; Markstrom, S.~L.; Hay, L.~E.; Viger, R.~J.; Norton, P.~A.; Driscoll, J.~M.; and LaFontaine, J.~H. 2018.
\newblock Description of the national hydrologic model for use with the precipitation-runoff modeling system (PRMS).
\newblock Technical Report 6-B9, US Geological Survey.

\bibitem[{Roberts et~al.(2018)Roberts, Cassula, Silveira, Bortoni, and Mendiburu}]{Roberts2018RobustHybridPower}
Roberts, J.~J.; Cassula, A.~M.; Silveira, J.~L.; Bortoni, E. d.~C.; and Mendiburu, A.~Z. 2018.
\newblock Robust multi-objective optimization of a renewable based hybrid power system.
\newblock \emph{Applied Energy}, 223: 52--68.

\bibitem[{Robinson et~al.(2019)Robinson, Hou, Malkin, Soobitsky, Czawlytko, Dilkina, and Jojic}]{Robinson2019LandCoverMapping}
Robinson, C.; Hou, L.; Malkin, K.; Soobitsky, R.; Czawlytko, J.; Dilkina, B.; and Jojic, N. 2019.
\newblock Large Scale High-Resolution Land Cover Mapping With Multi-Resolution Data.
\newblock In \emph{Proceedings of the IEEE/CVF Conference on Computer Vision and Pattern Recognition}.

\bibitem[{Sener and Koltun(2018)}]{Ozan2018MultiTaskLearning}
Sener, O.; and Koltun, V. 2018.
\newblock Multi-task learning as multi-objective optimization.
\newblock In \emph{Advances in Neural Information Processing Systems}, volume~31.

\bibitem[{Sun et~al.(2021)Sun, Jiang, Mudunuru, and Chen}]{sun2021explore}
Sun, A.~Y.; Jiang, P.; Mudunuru, M.~K.; and Chen, X. 2021.
\newblock Explore spatio-temporal learning of large sample hydrology using graph neural networks.
\newblock \emph{Water Resources Research}, 57(12): e2021WR030394.

\bibitem[{Suzuki et~al.(2020)}]{Suzuki2020MultiObjectiveBO}
Suzuki, S.; et~al. 2020.
\newblock Multi-objective Bayesian optimization using Pareto-frontier entropy.
\newblock In \emph{International Conference on Machine Learning}. PMLR.

\bibitem[{Tabari et~al.(2021)Tabari, Paz, Buekenhout, and Willems}]{Tabari2021sta}
Tabari, H.; Paz, S.~M.; Buekenhout, D.; and Willems, P. 2021.
\newblock Comparison of statistical downscaling methods for climate change impact analysis on precipitation-driven drought.
\newblock \emph{Hydrology and Earth System Sciences}, 25(6): 3493--3517.

\bibitem[{Terry et~al.(2022)Terry, Briggs, Kushner, Dickerson, Baldwin, Trottier, Haynes, Besteder, Glas, Doctor, Gazoorian, Odom, and Benton}]{Terry2022StreamTemperatureData}
Terry, N.; Briggs, M.; Kushner, D.; Dickerson, H.; Baldwin, A.; Trottier, B.; Haynes, A.; Besteder, C.; Glas, R.; Doctor, D.; Gazoorian, C.; Odom, W.; and Benton, J. 2022.
\newblock Stream temperature, dissolved radon, and stable water isotope data collected along headwater streams in the upper Neversink River watershed, NY, USA.

\bibitem[{Topp et~al.(2023)Topp, Barclay, Diaz, Sun, Jia, Lu, Sadler, and Appling}]{topp2023stream}
Topp, S.~N.; Barclay, J.; Diaz, J.; Sun, A.~Y.; Jia, X.; Lu, D.; Sadler, J.~M.; and Appling, A.~P. 2023.
\newblock Stream temperature prediction in a shifting environment: Explaining the influence of deep learning architecture.
\newblock \emph{Water Resources Research}, 59(4): e2022WR033880.

\bibitem[{{U.S. Environmental Protection Agency} and {U.S. Geological Survey}(2012)}]{EPA2012NHDPlus}
{U.S. Environmental Protection Agency}; and {U.S. Geological Survey}. 2012.
\newblock National {Hydrography} {Dataset} {Plus} ({NHDPlus}).

\bibitem[{{U.S. Geological Survey}(2019)}]{USGS2019NHD}
{U.S. Geological Survey}. 2019.
\newblock National Hydrography Dataset (ver. 2.1).

\bibitem[{{U.S. Geological Survey} et~al.(2023){U.S. Geological Survey}, {U.S. Department of Agriculture - Natural Resource Conservation Service}, {U.S. Environmental Protection Agency}, and {other Federal, State, and Local partners}}]{WBD2023}
{U.S. Geological Survey}; {U.S. Department of Agriculture - Natural Resource Conservation Service}; {U.S. Environmental Protection Agency}; and {other Federal, State, and Local partners}. 2023.
\newblock Watershed {Boundary} {Dataset} - {National}.

\bibitem[{Vaswani et~al.(2017)Vaswani, Shazeer, Parmar, Uszkoreit, Jones, Gomez, Kaiser, and Polosukhin}]{vaswani2017attention}
Vaswani, A.; Shazeer, N.; Parmar, N.; Uszkoreit, J.; Jones, L.; Gomez, A.~N.; Kaiser, L.; and Polosukhin, I. 2017.
\newblock Attention is all you need.
\newblock \emph{Advances in neural information processing systems}, 30.

\bibitem[{Ward(1985)}]{Ward1985ThermalCharacteristics}
Ward, J.~V. 1985.
\newblock Thermal characteristics of running waters.
\newblock In \emph{Perspectives in Southern Hemisphere Limnology: Proceedings of a Symposium}. Springer Netherlands.

\bibitem[{Williamson et~al.(2015)Williamson, Lant, Claggett, Nystrom, Milly, Nelson, Hoffman, Colarullo, and Fischer}]{williamson2015summary}
Williamson, T.~N.; Lant, J.~G.; Claggett, P.; Nystrom, E.~A.; Milly, P.~C.; Nelson, H.~L.; Hoffman, S.~A.; Colarullo, S.~J.; and Fischer, J.~M. 2015.
\newblock Summary of hydrologic modeling for the Delaware River Basin using the Water Availability Tool for Environmental Resources (WATER).
\newblock Technical Report 2015-5143, U.S. Geological Survey Scientific Investigations Report.

\bibitem[{Yousefi, Smith, and {\'{A}}lvarez(2019)}]{Yousefi2019aggregate}
Yousefi, F.; Smith, M.~T.; and {\'{A}}lvarez, M.~A. 2019.
\newblock {Multi-task learning for aggregated data using Gaussian processes}.
\newblock In \emph{Advances in Neural Information Processing Systems}.

\bibitem[{Yu, Yin, and Zhu(2018)}]{Yu2018STGCNAI}
Yu, B.; Yin, H.; and Zhu, Z. 2018.
\newblock Spatio-temporal graph convolutional networks: A deep learning framework for traffic forecasting.
\newblock In \emph{Proceedings of the 27th International Joint Conference on Artificial Intelligence}, 3634--3640.

\bibitem[{Zhang et~al.(2021)}]{Zhang2021Multibranch}
Zhang, F.; et~al. 2021.
\newblock Multi-branch and multi-scale attention learning for fine-grained visual categorization.
\newblock In \emph{MultiMedia Modeling: 27th International Conference, MMM 2021}, volume~27. Springer International Publishing.

\bibitem[{Zhang et~al.(2020)Zhang, Charoenphakdee, Wu, and Sugiyama}]{zhang2020aggregate}
Zhang, Y.; Charoenphakdee, N.; Wu, Z.; and Sugiyama, M. 2020.
\newblock Learning from Aggregate Observations.
\newblock In \emph{Advances in Neural Information Processing Systems}.

\bibitem[{Zhao, Liu, and Wang(2021)}]{Zhao2021FacialExpressionRecognition}
Zhao, Z.; Liu, Q.; and Wang, S. 2021.
\newblock Learning deep global multi-scale and local attention features for facial expression recognition in the wild.
\newblock \emph{IEEE Transactions on Image Processing}, 30: 6544--6556.

\bibitem[{Zhou et~al.(2018)}]{Zhou2018MultiObjectiveRobust}
Zhou, Q.; et~al. 2018.
\newblock A multi-objective robust optimization approach based on Gaussian process model.
\newblock \emph{Structural and Multidisciplinary Optimization}, 213--233.

\end{thebibliography}
